\documentclass[a4paper]{article}

\usepackage[english]{babel}
\usepackage[utf8x]{inputenc}
\usepackage[T1]{fontenc}

\usepackage[a4paper,top=3cm,bottom=2cm,left=3cm,right=3cm,marginparwidth=1.75cm]{geometry}

\usepackage{amsmath}
\usepackage{amsthm}
\usepackage{graphicx}
\usepackage{caption}
\usepackage[]{algorithm2e}
\usepackage[colorinlistoftodos]{todonotes}
\usepackage[colorlinks=true, allcolors=blue]{hyperref}
\usepackage{authblk}
\usepackage{hhline}
\usepackage[titletoc]{appendix}

\newcommand{\abcdresults}[7]{#1 & #3 & #4 & #7 & #5 & #6 & #2\\}
\usepackage{makecell}
\usepackage[section]{placeins}

\usepackage{tcolorbox}

\DeclareMathOperator*{\argmax}{arg\,max}
\DeclareMathOperator*{\argmin}{arg\,min}
\newtheorem{theorem}{Theorem}

\makeatletter
\renewcommand*{\@fnsymbol}[1]{\ensuremath{\ifcase#1\or \dagger\or \ddagger\or
    \mathsection\or \mathparagraph\or \|\or **\or \dagger\dagger
    \or \ddagger\ddagger \else\@ctrerr\fi}}
\makeatother

\title{Adaptive Ensemble of Classifiers with Regularization for Imbalanced Data Classification}
\author[1,2]{Chen Wang\thanks{Work was done at Sichuan University. The author is now at Rutgers University -- New Brunswick.}}
\author[2]{Chengyuan Deng}
\author[3]{Zhoulu Yu}
\author[4]{Dafeng Hui}
\author[1]{Xiaofeng Gong}
\author[1,*]{Ruisen Luo}
\affil[1]{Sichuan University, Chengdu, China 610065}
\affil[2]{Rutgers University -- New Brunswick, Piscataway, USA 08854}
\affil[3]{Zhejiang University, Hangzhou, China 310058}
\affil[4]{Tennessee State University, Nashville, USA 37209}

\affil[*]{Corresponds to Ruisen Luo: \emph{rsluo@scu.edu.cn}}

\begin{document}
\date{}
\maketitle

\begin{abstract}


The dynamic ensemble selection of classifiers is an effective approach for processing label-imbalanced data classifications. However, such a technique is prone to overfitting, owing to the lack of regularization methods and the dependence of the aforementioned technique on local geometry. In this study, focusing on binary imbalanced data classification, a novel dynamic ensemble method, namely adaptive ensemble of classifiers with regularization (AER), is proposed, to overcome the stated limitations. The method solves the overfitting problem through implicit regularization. Specifically, it leverages the properties of stochastic gradient descent to obtain the solution with the minimum norm, thereby achieving regularization; furthermore, it interpolates the ensemble weights by exploiting the global geometry of data to further prevent overfitting. According to our theoretical proofs, the seemingly complicated AER paradigm, in addition to its regularization capabilities, can actually reduce the asymptotic time and memory complexities of several other algorithms. We evaluate the proposed AER method on seven benchmark imbalanced datasets from the UCI machine learning repository and one artificially generated GMM-based dataset with five variations. The results show that the proposed algorithm outperforms the major existing algorithms based on multiple metrics in most cases, and two hypothesis tests (McNemar's and Wilcoxon tests) verify the statistical significance further. In addition, the proposed method has other preferred properties such as special advantages in dealing with highly imbalanced data, and it pioneers the research on the regularization for dynamic ensemble methods.


\end{abstract}

Keywords - Adaptive ensemble; Gradient Boosting Machines; Regularization; Imbalanced data classification 

\section{Introduction}
\label{IntroductionSection}
Imbalanced data classification refers to the classification of datasets with significantly different instance numbers across classes \cite{haixiang2017learning}. Specifically, for the binary imbalanced data classification problem, there is usually a dominating number of instances from one class (the majority class) and a few instances belonging to the other class (the minority class). The problem of binary imbalanced data classification is common in engineering and scientific practices  \cite{huda2016BrainTumorImbalancedData,ashkezari2013EnergySectorApplication,triguero2015rosefw}. The problem is non-trivial, because most of the general-purpose classification methods will overwhelmingly favor the majority class in the label-imbalanced scenario, leading to significant performance degradation. Consequently, the development of binary imbalanced classification algorithms has become an independent and active research area. \par

Among the popular algorithms for binary imbalanced classification, the dynamic ensemble of classifiers has attracted significant attention. It works by training multiple classifiers using different subsets of the data, and \emph{dynamically} selecting from, or combining them, during the inference process. By picking the most competent classifier(s) for \emph{each} specific test instance, this approach can mitigate the “majority favoritism” in imbalanced data classification \cite{cruz2018MulticlassDynamicEnsembleSurvey,roy2018study}. There are various advanced algorithms built on the strategy of the dynamic ensemble model, and the novelty of most of them lies in their use of certain new techniques to `ensemble' models. For instance, \cite{costa2018combining} proposes a generalized mixture function to combine different classifiers, and \cite{brun2018framework} proposes an adaptive ensemble method based on the classification problem. We review a few similar methods in this paper, and more details are presented in Section \ref{LiteratureReviewSection}. \par


Despite the success of the dynamic ensemble of classifiers with regard to various tasks, we are unaware of any existing model that addresses the \emph{overfitting} exhibited by such classifiers. Overfitting is a common problem, wherein the behavior of the classifiers overly fit the training data. This adversely affects the performance on the test data, because not all the information in the training data is useful (e.g. noises). At first glance, it appears that the dynamic ensemble of classifiers can safely circumvent the curse of overfitting, because they utilize the test data during the selection of classifiers. However, because each classifier is usually trained using a small subset of data (which contains information from the local geometry only), dynamically picking \emph{the most competent } of them can lead to the overfitting of the local geometry of these classifiers. Even if we interpolate the dynamic ensemble with a set of (fixed) trained weights for the classifiers, the overfitting problem will persist, as the weights are obtained purely from the training data. Hence, it seems there is no simple solution to the overfitting problem of the dynamic ensemble of classifiers. \par

We solve the aforementioned problems using the regularization effect arising from gaussian mixture model (GMM)-based resampling and the stochastic gradient descent (SGD) algorithm. The proposed method is called the adaptive ensemble of classifiers with regularization (AER), where the term “with regularization” refers to the two types of regularization schemes that are developed in this study. The AER method first performs data resampling based on the GMM \cite{barber2012TextMoG,reynolds1995robust}. We will generate two types of subsets. The first type will have a broader inclusion of the points from the majority class, and the second will have an almost balanced number of instances from the two classes. The former type of subsets can force the classifiers to consider the global geometry; therefore, this is the regarded as the \emph{first regularization} to alleviate the overfitting problem. The latter type of subsets provides information on the local geometries to ensure they fit sufficiently powerful classifiers. After completing the resampling process, one individual classifier is learned for each sampled subset, and we explicitly learn a set of fixed coefficients/weights by optimizing the cross-entropy loss of the combined model with the SGD. The adaptation of the SGD is the \emph{second regularization}, and its effectiveness has been verified by numerous studies \cite{gardner1984learning,zhang2016implicitRegularisationSGD,schmidhuber2015SGD,bottou2018optimization}. During the inference procedure, the normalized coefficient of each individual classifier will be determined by a combination of the on-the-fly likelihood and the trained classifier coefficients. \par

We theoretically and empirically evaluate the performance of the proposed AER. From a theoretical perspective, we analyze the time and space complexity of the AER model, and prove that the seemingly complicated AER model actually requires less time and memory to train. From an empirical perspective, we test the performance of the AER model, using the XGBoost classifier \cite{Chen2016XGBoost} (we refer to the combined method as \emph{AER-XGBoost}) based on seven imbalanced UCI machine learning datasets and a GMM-generated dataset with five variations. Based on multiple metrics, experimental results reveal that the AER-XGBoost model exhibits competitive performances, outperforming multiple standard methods, such as the SVM and decision tree, and state-of-the-art methods, such as the focal loss neural network \cite{lin2018FocalLoss}, vanilla XGBoost \cite{Chen2016XGBoost}, focal loss XGBoost \cite{wang2019imbalance}, and the LightGBM model \cite{ke2017lightgbm}. The Mcnemar's and Wilcoxon signed-rank tests are performed to further validate the superior performance of the AER, and the results are mostly sufficient to reject the null hypothesis for performance difference. We note that the AER generally performs significantly better in severe label-imbalanced and complex decision boundary scenarios.

The rest of the paper is structured as follows: Section \ref{MethodSection} introduces the algorithm in detail, with its properties. Section \ref{ComplexitySection} analyzes the advantageous time and memory complexity of the proposed algorithm. Experimental Framework and and results analysis are demonstrated in Section \ref{ExperimentFramework} and Section \ref{ExperimentSection} respectively, and related discussions are presented in Section \ref{sec:discussion}. Lastly, Section \ref{ConclusionSection} provides a general conclusion of the paper.

\section{Related Work}
\label{LiteratureReviewSection}
Imbalanced data classification refers to the classification problem where the number of samples for each class label is not balanced, or, where the class distribution is biased or skewed \cite{haixiang2017learning}. Since most of the standard classifiers assume relatively balanced class distributions and equal mis-classification costs, the class-imbalance can be perceived as a form of data irregularity \cite{das2018handling}, and it could significantly deteriorate the performances of classifiers. Performing high-accuracy classification using imbalanced data has been a challenge for a long time, and there have been considerable number of publications discussing novel methods to address the problem. The methods can be roughly categorized into four branches \cite{fernandez2013analysingImbalancedMethods,krawczyk2016ImbalancedLearningChallenge}.  The first branch is the re-sampling methods that generate balanced data by under-sampling the majority class, and/or over-sampling the minority instances \cite{more2016ResamlpingSurvey}. The second branch is the cost-sensitive algorithms that address the problem by using imbalance-sensitive target functions, and assigning special loss functions explicitly (\cite{khan2017costSensitiveTarget}) or implicitly (\cite{datta2015near,raghuwanshi2018class,shukla2019online}). The third branch is the one-class learning methods that solve the label-imbalance problem by learning the representation of the majority/minority data \cite{bellinger2012oneClassLearning,luo2018feature}. The final branch is the ensemble methods that contain the dynamic ensemble of classifiers, as employed in this study. \par

The idea of the ensemble methods (and, by implication, the dynamic ensemble of classifiers) is to utilize `the wisdom of the many'; a typical ensemble method will fit multiple classifiers and/or re-sampled multiple datasets, and then combine them in some way to obtain final predictions \cite{de2005BaggingAndBoosing,galar2011LogisticRegressionEnsemble,khoshgoftaar2007RandomForestImbalancedData,wang2017novelSVMensemble}. On the other hand, the static ensemble of classifiers approach will complete the training of the individual classifiers and their weights based solely on the training data \cite{galar2012ReviewEnsembleImbalancedData}. This approach requires less computational resource, as the predictions can be obtained using a fixed model. However, when encountering noises in the minority or complex and rapidly changing decision boundary between the classes, the static ensemble methods often suffer performance degradation \cite{wozniak2014SurveyOnHybridIntelligentSystem,galar2012ReviewEnsembleImbalancedData,khoshgoftaar2011ComparingBaggingAndBoosting}. On the other hand, the dynamic ensemble methods will change the combination of classifiers on-the-fly, according to the test instances. Compared with its static counterpart, the dynamic ensemble of classifiers introduces greater model flexibility and representation power. Multiple studies have empirically shown that the dynamic ensemble is a more preferred choice, based on different metrics \cite{cruz2018MulticlassDynamicEnsembleSurvey,roy2018study,Lima2014Improving}. \par

The dynamic ensemble approach can be roughly divided into two categories: the dynamic classifier selection (DCS) approach, wherein the most competent classifier is selected dynamically (\cite{woods1997EearlyDynamicEnsemble}), and the dynamic ensemble selection (DES), wherein multiple classifiers are selected for prediction (\cite{ko2008DCSandDES}). Earlier works in the area often adopt the DCS strategy; for instance, Woods (1997) \cite{woods1997EearlyDynamicEnsemble} utilizes a “rank-based” method to select the most competitive classifier. Conversely, because the DES is more flexible, modern scientists have investigated more novel methods based on it. Most of the recent models pin their novelty on the renewed strategies to combine/ensemble the classifiers. For instance, Lin et al. (2014) \cite{lin2014libd3cClusterBased} proposed a method to ensemble classifiers based on clustering results; Cruz et al. (2015) \cite{cruz2015MetaLearningEnsemble} designed an algorithm to combine classifiers using meta-learning; Xiao et al. (2012) \cite{xiao2012DynamicWithCostSensitive} introduced cost-sensitive criteria to determine the weights of the ensemble; Krawczyk et al. (2018) \cite{krawczyk2018dynamic} used the dynamic ensemble of one-class classifiers to train the model with regard to multiple classes; and Brun et al., (2018) \cite{brun2018framework} proposed adjusting the ensemble based on the difficulty of classification. Nevertheless, despite the fruitful results in the literature, to the best of our knowledge, we do not know any method under the DES framework wherein regularization is introduced (to avoid overfitting). \par

We assume that the absence of such research is because there is no obvious approach to introduce regularization to the dynamic ensemble model. Conventional norm-based regularization works by minimizing the norms of the weights; however, in the dynamic ensemble scenario, the weight of each classifier cannot be shrunk, because they should be in the range of $[0,1]$, and sum up to $1$ (a simplex). Other existing regularization techniques are either not applicable to the scenario (e.g., NoiseOut \cite{babaeizadeh2016noiseout}, which is designed for neural networks), or considered too “aggressive” for linear combination (e.g., dropout \cite{srivastava2014dropout}, which will block some classifiers entirely). To overcome the above obstacles, the AER is regularized implicitly based on two techniques: global-geometry interpolation and the SGD algorithm. The regularization effect of the first approach is derived from previous ideas on data resampling, and the central idea is to reduce the impact of the noise from the overlap between the classes \cite{liu2008exploratory,lee2018overlap}. The regularization power of the second approach (SGD) is supported by its attribute of converging to a solution with a minimum $\ell_{2}$ norm \cite{lin2015iterative}, and multiple theoretical and empirical results have proved its effectiveness \cite{zhang2016implicitRegularisationSGD,Lin2016Generalization}. \par 


Binary imbalanced classification has a broad range of applications. In bioscience and medical research, imbalanced data classification has been utilized to identify tumor \cite{huda2016BrainTumorImbalancedData}, and diagnose cancer \cite{krawczyk2016CancerImbalanceData}. Likewise, in software engineering, it has been employed to detect bugs \cite{xia2015BugDetection} or malignant software \cite{chen2018MalwareDetection}. In other fields, such as financial fraud detection \cite{mardani2013FraudDetection} and power transformation \cite{ashkezari2013EnergySectorApplication}, imbalanced data classifications are also comprehensively employed. Guo (2017) \cite{haixiang2017learning} surveys the applications of imbalanced data classification, and shows the promising potential in applying such a technique to a broader range of problems. \par

\section{Methods}
\label{MethodSection}
\begin{figure}
\centering
\includegraphics[width=1.0\textwidth]{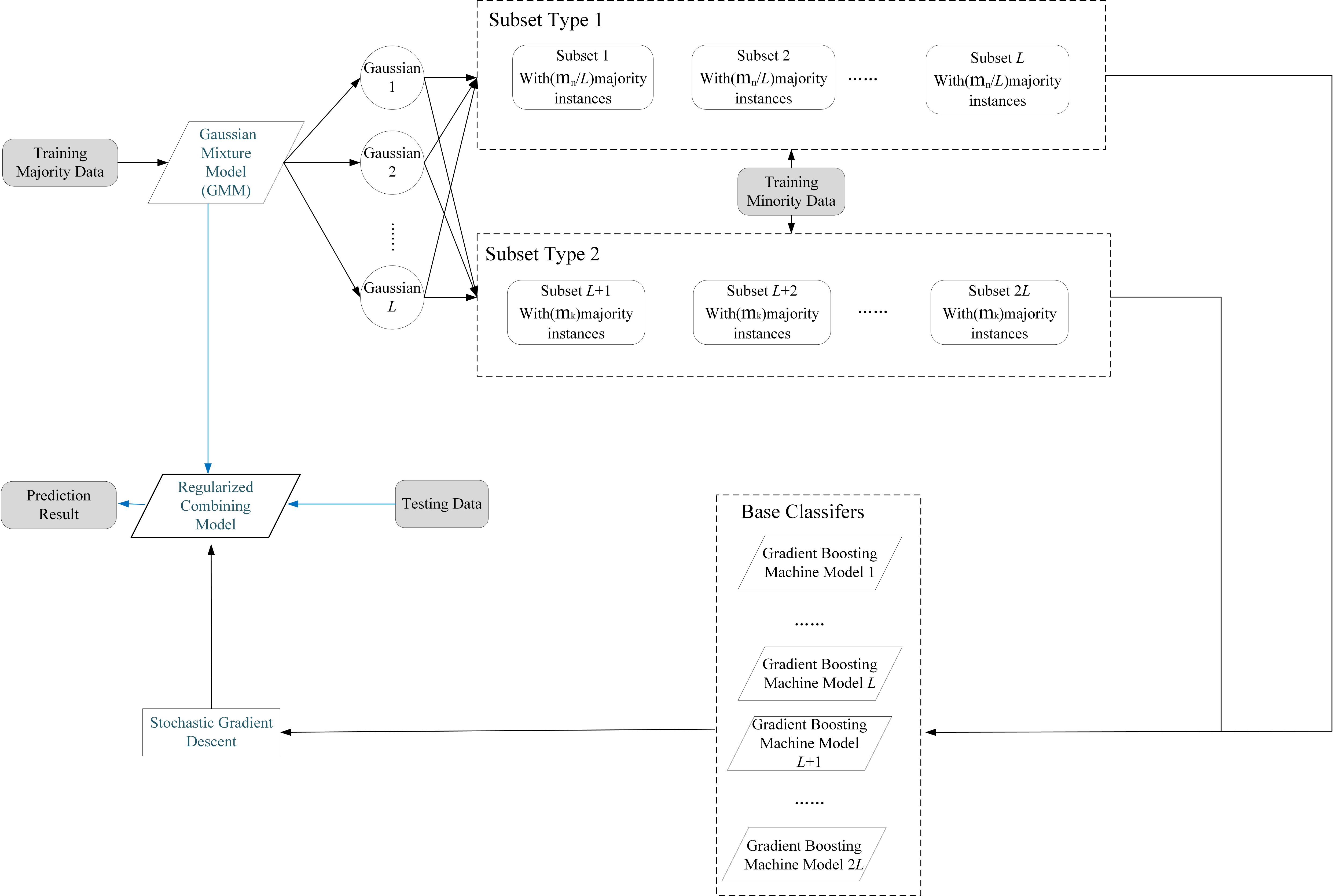}
\caption{\label{fig:overall}The overall process of the proposed AER algorithm}
\end{figure}

In this section, we introduce the details of the proposed AER model. The structure is laid out as follows: Section \ref{subsec:GMMandDataGeneration} will introduce the GMM-fitting and generation of the two types of subsets; Section \ref{subsec:XGBoostFitting} will discuss the specific implementation with XGBoost, which is the individual `base' classifiers used in the experiments; the SGD training for the ensemble of classifiers will be illustrated in Section \ref{subsec:SGDtrainEnsemble}; and finally, the weight interpolation/combination and probabilistic prediction will be shown in Section \ref{subsec:WeightInterpolation}. The overall procedure of the algorithm is shown in Figure \ref{fig:overall}. \par
Before diving into the details, we shall specify the notations that will be used. We use $\boldsymbol{x}$ to denote a single instance of the data in the dataset $\boldsymbol{X}$. To distinguish the majority and minority data, we use $\boldsymbol{X}_{k}$ (\textbf{k}ey data, usually the minority) to denote the set of minority data and $\boldsymbol{X}_{n}$ (\textbf{n}on-key data, usually the majority) to represent the set of majority data. The size of the dataset is denoted by $m$, and the dimension (the number of features) is represented by $n$; we use $m_{k}$ and $m_{n}$ to denote the number of points of the minority and the majority data. Finally, $L$ denotes the number of Gaussian distributions in the GMM, and $2L$ is the number of classifiers, according to our setup.
\subsection{Gaussian Mixture Model Fitting and Subset Generation}
\label{subsec:GMMandDataGeneration}
The GMM is a popular model in unsupervised learning and data manifold representation. The basic idea behind the GMM is straightforward; it utilizes the modeling capability of the Gaussian distribution, and extends it to multiple centroids to improve the expressiveness. The likelihood of a single instance in the GMM can be denoted as follows:\\
\begin{equation}
\label{MoGbaiscEquation}
p(\boldsymbol{x}|\{\boldsymbol{\mu}_{l},\boldsymbol{\Sigma}_{l},w_{l}\}) = \sum_{l=1}^{L} w_{l}\mathcal{N}(\boldsymbol{x}|\boldsymbol{\mu}_{l},\boldsymbol{\Sigma}_{l})
\end{equation}
where $\mathcal{N}(\cdot|\boldsymbol{\mu},\boldsymbol{\Sigma})$ denotes a multivariate Gaussian distribution, with $\boldsymbol{\mu}$ as the mean and $\boldsymbol{\Sigma}$ as the co-variance. When fitting the model, the parameters can be obtained by maximizing the log-likelihood:\\
\begin{equation}
\label{MoGfittingMLE}
\begin{aligned}
\{\hat{\boldsymbol{\mu}}_{l},\hat{\boldsymbol{\Sigma}}_{l},\hat{w}_{l}\} & = \argmax_{\boldsymbol{\mu}_{l},\boldsymbol{\Sigma}_{l},w_{l}} \log(\prod_{i=1}^{m} \sum_{l=1}^{L} w_{l}\mathcal{N}(\boldsymbol{x}_{i}|\boldsymbol{\mu}_{l},\boldsymbol{\Sigma}_{l})) \\
& =  \argmax_{\boldsymbol{\mu}_{l},\boldsymbol{\Sigma}_{l},w_{l}} \sum_{i=1}^{m}\log(\sum_{l=1}^{L} w_{l}\mathcal{N}(\boldsymbol{x}_{i}|\boldsymbol{\mu}_{l},\boldsymbol{\Sigma}_{l}))
\end{aligned}
\end{equation}
Equation \ref{MoGfittingMLE} can be solved using an expectation-maximization (E-M) algorithm with a super-linear convergence rate \cite{xu1996EMalgorithmSuperLinearConvergence}. In our program, the package of the GMM provided by SciKit-learn \cite{pedregosa2011scikit} is directly adopted, to perform the fitting procedure of the GMM. \par

It may be observed that the GMM is sensitive to initialization. To obtain stable results, the optimization will be performed $5$ times for each training procedure, and the model with the highest log-likelihood will be selected. Another non-learnable parameter in the GMM is the number of Gaussian distributions. In the proposed algorithm, this hyper-parameter also determines the number of components in the final ensemble. To obtain the optimal number of components that optimally balances the likelihood and computational complexity, the Bayesian information criterion (BIC) is adopted. The BIC metric can be computed as follows: \\
\begin{equation}
\label{equ:BIC_computation}
\text{BIC} = \log(m)\|\boldsymbol{f}\|-2\sum_{i=1}^{m}\log p(\boldsymbol{x}_{i})
\end{equation}
where $\|\boldsymbol{f}\|$ stands for the number of the parameters of the model, and $p(\boldsymbol{x}_{i})$ indicates the likelihood of the instance $\boldsymbol{x}_{i}$. In terms of the GMM presented in our method, the BIC will be computed as follows:\\
\begin{equation}
\label{equ:BIC_computation_MoG}
\text{BIC} = \log(m)\|\boldsymbol{f}\|-2\sum_{i=1}^{m}\log(\sum_{l=1}^{L} w_{l}\mathcal{N}(\boldsymbol{x}_{i}|\boldsymbol{\mu}_{l},\boldsymbol{\Sigma}_{l})) 
\end{equation}
 In the training process, a “pool” of the possible numbers of Gaussian centroids will be given, and the algorithm will compute the BIC of each model, and pick the one with the least BIC. We choose the BIC over the Akaike information criterion (AIC), because the BIC tends to favor the model that overfits the data less \cite{dziak2012AICBICmeasure}. Because regularization plays an important role in our algorithm, the BIC is adopted as the choice of hyper-parameter. \par
After obtaining the GMM with $L$ Gaussian distributions, we form $2L$ sub-datasets based on two types of schemes. The first type of scheme will be selecting $\lfloor \frac{m_{n}}{L} \rfloor$ most representative data from each of the Gaussian distribution. Specifically, for each Gaussian distribution $\mathcal{N}(\boldsymbol{x}_{i}|\boldsymbol{\mu}_{l},\boldsymbol{\Sigma}_{l})$, the algorithm will select $\lfloor \frac{m_{n}}{L} \rfloor$ instances with the highest log-likelihood. For the second type of paradigm, the algorithm will generate $L$ subsets with $m_{l}$ majority instances selected based on the highest likelihoods, with respect to each Gaussian component, and concatenate it with the $\lfloor0.5*m_{l}\rfloor$ majority instances randomly selected from the set. Both of the generated majority datasets will be combined with all the minority instances.\par

After obtaining the above data, a Tomek link \cite{tomek1976Tlink} will be used to remove instances considered as noises from the first type of data subsets. Tomek links are based on the idea that if two instances are mutually nearest neighbors that belong to different classes, they would be 'overlapping' instances between classes, and be likely to be noises. Formally, for two given data instances $\{\boldsymbol{x}_{i},\boldsymbol{x}_{j} \} \in \boldsymbol{X}$ and given distance measure $d(\cdot,\cdot)$, if for any $\boldsymbol{x}_{k} \in \boldsymbol{X}$, the following exists:\\
\begin{equation}
\label{TomekLinkCondition}
\begin{cases}
d(\boldsymbol{x}_{i},\boldsymbol{x}_{j})<d(\boldsymbol{x}_{i},\boldsymbol{x}_{k}) \\
d(\boldsymbol{x}_{i},\boldsymbol{x}_{j})<d(\boldsymbol{x}_{j},\boldsymbol{x}_{k})
\end{cases}
\end{equation}
Then, $\{\boldsymbol{x}_{i},\boldsymbol{x}_{j} \}$ will be considered a pair of Tomek link. If the corresponding labels $\{y_{i}, y_{j}\}$ of the Tomek link pair belong to different classes, then we regard the majority and/or minority instance in the pair as noise, and remove one or both of them. In our algorithm, because the minority will be the more important part to be spotted, the majority instances in the Tomek links will be removed.\par

By performing the above process, there will be $2L$ available sub-datasets, with less significant skewness of labels. As we have explained above, the reason for adopting a combination of two type of re-sampling methods is that this strategy can separately achieve two different goals satisfactorily. The first type of subset can preserve the information of the global geometry, and contribute to the recognition of the majority instances; this helps the AER to avoid overfitting on the overlapping parts between the majority and the minority instances (i.e., a `zoom-out perspective' to `pull back' the decision boundary from the complexity overlapping part between the two classes). The second type of subset emphasizes the local geometry, and improves the accuracy of spotting minority instances (i.e. a “zoomed-in perspective” to focus on learning the complex decision boundary on the overlapping part). Furthermore, because the second type of datasets focus \emph{only} on the local geometry, we will concatenate them with $\lfloor0.5*m_{k}\rfloor$ majority instances randomly selected to prevent the decision boundary that is away from the local part from “going wild.”\par
The overall procedure of the GMM-fitting and subset generation is depicted as Algorithm \ref{alg:MixtureOfGaussianAndSubsampling}.\\
\begin{algorithm}[H]
\KwData{majority data $\boldsymbol{X}_{n}$, minority data $\boldsymbol{X}_{k}$, candidate numbers for $L$}
\KwResult{Gaussian Mixture Model $\{w_{l},\boldsymbol{\mu}_{l},\boldsymbol{\Sigma}_{l}\}_{l=1:L}$, $2L$ balanced datasets}
\For{$L_i$ \textbf{in} candidate $L$ values}{
  \For{$trial\gets 1$ \KwTo $5$}{
      initialize Gaussian Mixture Model parameters\;
      optimize Gaussian Mixture Model with equation \ref{MoGfittingMLE} via E-M algorithm\;
  }
  Retrieve the model with the highest log-likelihood\;
  Compute corresponding BIC using equation \ref{equ:BIC_computation_MoG}\;
}
Select the model with the smallest BIC value\;
\For{$i\gets 1$ \KwTo $L$}{
  select $\lfloor \frac{m_{n}}{L} \rfloor$ most relevant samples from $\boldsymbol{X}_{n}$, according to the log-likelihood of the Gaussian centroid $i$\;
    combine the obtained data with $\boldsymbol{X}_{k}$ to form the first type of subset $\boldsymbol{\hat{X}}_{k}$\;
    select $m_{k}$ most relevant samples from $\boldsymbol{X}_{n}$, according to the log-likelihood of Gaussian centroid $i$ and combine them with $\lfloor 0.5*m_{k} \rfloor$ randomly selected majority instances\;
  combine the obtained data with $\boldsymbol{X}_{k}$ to form the second type of subset $\boldsymbol{\check{X}}_{k}$\;
  perform a Tomek link removal for $\boldsymbol{\hat{X}}_{k}$ with equation provided as the condition \ref{TomekLinkCondition}\;
}
\caption{Gaussian Mixture Model Fitting and Balanced Data Generating}
\label{alg:MixtureOfGaussianAndSubsampling}
\end{algorithm}
\subsection{Fitting of Individual Base Classifier }
\label{subsec:XGBoostFitting}
As has been stated above, the specific classifier implemented in this study is the gradient boosting machine (GBM) (\cite{friedman2001greedy}), a boosting-based algorithm. The GBM model can be expressed as follows:\\
\begin{equation}
G_{T}(\boldsymbol{X};\{\boldsymbol{\theta}_{t}\}_{t=1,2,..,T}) = \sum_{t=1}^{T} \alpha_{t}g_{t}(\boldsymbol{x};\boldsymbol{\theta}_{t})
\end{equation}
Similar to other boosting methods, the training strategy of the GBM is to learn from previous mistakes. Specifically, the individual sub-model of the GBM at the $T$-th step will set the gradient of the loss function, with respect to the model, up to $(T-1)$-th step as the current `labels', which can be expressed as follows:\\
\begin{equation}
\label{PseudoResidualComputation}
r_{T-1} = -\frac{\partial L(\boldsymbol{y},G_{T-1}(\boldsymbol{X};\{\boldsymbol{\theta}_{t}\}_{t=1,2,..,T-1}))}{\partial G_{T-1}(\boldsymbol{X};\{\boldsymbol{\theta}_{t}\}_{t=1,2,..,T-1})}
\end{equation}
where $L(\cdot,\cdot)$ is used to denote any kind of loss function, and is usually the square loss for regression and the cross-entropy loss for classification. The gradient computed with equation \ref{PseudoResidualComputation} is also named the `pseudo-residual'; because the gradient will be calculated at each step, the overall method is tagged the GBM. After obtaining the current target, the parameter of the sub-model at the $T$-th step can be denoted as follows:\\
\begin{equation}
\label{equ:KthStepThetaFitting}
\boldsymbol{\theta}_{T} = \argmin_{\boldsymbol{\theta}}L(r_{T-1},G(\boldsymbol{X};\boldsymbol{\theta}))
\end{equation}
The overall model at the $T$-th step is further determined by a 'learning rate' $\alpha_{T}$ that can be obtained by optimizing the following target function:\\
\begin{equation}
\label{equ:alphaValueOptimization}
\alpha_{T} = \argmin_{\alpha}L(\boldsymbol{y},G_{T-1}(\boldsymbol{X};\{\boldsymbol{\theta}_{t}\}_{t=1,2,..,T-1})+\alpha g(\boldsymbol{X};\boldsymbol{\theta}_{T}))
\end{equation}
The above optimization can be simply solved, either by taking partial derivative or through a line search. By iterating the procedure from Equation \ref{PseudoResidualComputation} to \ref{equ:alphaValueOptimization}, until it matches the convergence criteria, the integrated GBM model will be obtained.\par
In our implementation, an integrated, high-efficient, and scalable gradient boosting interface, XGBoost \cite{Chen2016XGBoost}, is employed to fit, and make prediction with the GBM. For each sub-set of data generated, the algorithm will fit one XGBoost model, denoting as $f_{l}(\cdot)$.
\subsection{Stochastic Gradient Descent Training for the Ensemble of Classifiers}
\label{subsec:SGDtrainEnsemble}
After the training with the individual models, each classifier will be able to give a class prediction (0 or 1) for every data instance. The next step is to train the combination of individual classifiers using the SGD algorithm. For convenience, each individual model will be denoted as $f_{l}(\cdot)$. We can denote the linear combination of the models as follows:\\
\begin{equation}
\label{equ:LinearCombinationGBMmodel}
\begin{aligned}
F(\boldsymbol{X}) &= \sum_{l=1}^{2L} w_{l} * f_{l}(\boldsymbol{X})\\
&\text{s.t. } 
\begin{aligned}
&\sum_{l=1}^{2L} w_{l} = 1\\
&0\leq w_{l} \leq 1, \quad \forall l \in \{1,2,...,2L\}
\end{aligned}
\end{aligned}
\end{equation}
The constraints of Equation \ref{equ:LinearCombinationGBMmodel} is to guarantee that the values of the predictions will be between $[0,1]$; thus, it could simply be transferred to a binary-class prediction. To train the model, the two-class cross-entropy loss is adopted:\\
\begin{equation}
\label{equ:CrossEntropyLossLinearCombination}
\begin{aligned}
L(\boldsymbol{y},\boldsymbol{X}) &= -\sum_{i=1}^{m}y_{i}\log(\sum_{l=1}^{2L} w_{l} \cdot f_{l}(\boldsymbol{x}_{i}))+(1-y_{i})\log(1-\sum_{l=1}^{2L} w_{l} \cdot f_{l}(\boldsymbol{x}_{i}))\\
& = -\sum_{i=1}^{m}y_{i}\log( \boldsymbol{w}^{T}\boldsymbol{f}(\boldsymbol{x}_{i}))+(1-y_{i})\log(1-\boldsymbol{w}^{T}\boldsymbol{f}(\boldsymbol{x}_{i}))
\end{aligned}
\end{equation}
where the second line of the above equation is the vectorized expression. Note that because there is a constraint on $\boldsymbol{w}$, the above optimization cannot be accomplished by simply taking derivatives and setting them to 0. Furthermore, to perform the SGD algorithm, the gradient of the target \ref{equ:CrossEntropyLossLinearCombination}, with respect to $\boldsymbol{w}$ should be as follows:\\
\begin{equation}
\label{equ:GradientOfLinearCombination}
\begin{aligned}
\frac{\partial L(\boldsymbol{y},\boldsymbol{X})}{\partial \boldsymbol{w}} & = -\sum_{i=1}^{m} (\frac{y_{i}}{\boldsymbol{w}^{T}\boldsymbol{f}(\boldsymbol{x}_{i})}-\frac{1-y_{i}}{(1-\boldsymbol{w}^{T}\boldsymbol{f}(\boldsymbol{x}_{i}))})\boldsymbol{f}(\boldsymbol{x}_{i})\\
& = \sum_{i=1}^{m} (\frac{1}{(1-y_{i})-\boldsymbol{w}^{T}\boldsymbol{f}(\boldsymbol{x}_{i})})\boldsymbol{f}(\boldsymbol{x}_{i})
\end{aligned}
\end{equation}
If gradient descent is applied, the gradient in equation \ref{equ:GradientOfLinearCombination} does not guarantee that the sum of $\boldsymbol{w}$ will be 1, neither does it warrant that each $w_{k}$ will be in the interval of $[0,1]$. We provide a simple remedy to this; we can simply re-normalize the weight after learning. In addition, for the weights exceeding the limit of the interval, we can re-scale them to the limit value (0 or 1). In this way, the update formula of $\boldsymbol{w}$ should be as follows:\\
\begin{equation}
\begin{aligned}
\label{WeightUpdateFormula}
\boldsymbol{w}_{t+1} & = \frac{S(\boldsymbol{w}_{t} - \gamma_{t}\nabla_{\boldsymbol{w}_{t}}L(\boldsymbol{y},\boldsymbol{X}))}{\sum_{l=1}^{L}S(\boldsymbol{w}_{t} - \gamma_{t}\nabla_{\boldsymbol{w}_{t}}L(\boldsymbol{y},\boldsymbol{X}))_{l}}\\
& = \frac{S(\boldsymbol{w}_{t} - \gamma_{t}\sum_{i}^{m} (\frac{1}{(1-y_{i})-\boldsymbol{w}_{t}^{T}\boldsymbol{f}(\boldsymbol{x}_{i})})\boldsymbol{f}(\boldsymbol{x}_{i}))}{\sum_{l=1}^{L}S(\boldsymbol{w}_{t} - \gamma_{t}\sum_{i}^{m} (\frac{1}{(1-y_{i})-\boldsymbol{w}_{t}^{T}\boldsymbol{f}(\boldsymbol{x}_{i})})\boldsymbol{f}(\boldsymbol{x}_{i}))_{l}}
\end{aligned}
\end{equation}
where $\gamma_{t}$ is the learning rate, and $S(\cdot)$ denotes the re-scaling(mapping) function floor at 0 and ceiling at 1. This can be mathematically denoted as follows:\\
\begin{equation}
\label{equ:SaturatedLinearFunction}
S(x) = \frac{x-x_{min}}{x_{max}-x_{min}}
\end{equation}
It is recommended that the learning rate should be set less than $\frac{1}{||\nabla_{\boldsymbol{w}_{t}}L(\boldsymbol{y},\boldsymbol{X})||}$, to ensure convergence. To determine whether the training procedure has converged, the relative change in the cross-entropy loss is adopted as the metric. \par
Another concern regarding using the SGD method is how to initialize the coefficients, as the optimization is sensitive to initial values. In the proposed algorithm, the initialization of parameters is accomplished by the combination of the AIC and BIC. Similar to the BIC, the AIC can be expressed as follows:\\
\begin{equation}
\label{equ:AIC_computation}
\text{AIC} = 2\|\boldsymbol{f}\| - 2\sum_{i=1}^{m}\log(p(\boldsymbol{x}_{i}))
\end{equation}
where $p(\boldsymbol{x}_{i})$ indicates the likelihood of instance $\boldsymbol{x}_{i}$. Combining equations \ref{equ:BIC_computation} and \ref{equ:AIC_computation}, we could compute a combined metric as follows:\\
\begin{equation}
\boldsymbol{\tilde{w}}_{l} = \lambda_{A}\cdot \text{AIC}_{l} + (1-\lambda_{A})\cdot \text{BIC}_{l}
\end{equation}
where $\lambda_{A}$ is the parameter for striking a balance between the AIC and BIC; experiments demonstrate that $\lambda_{A} = 0.6$ could be an effective trade-off. Here, the AIC and BIC are computed with respect to each classifier, and lower AIC/BIC values indicate a more credible solution. Thus, we can use the normalized reciprocal of $\boldsymbol{\tilde{w}}_{l}$ values to initialize the linear combination. The initial values can be denoted as follows:\\
\begin{equation}
\label{InitializationOfCombinationWeights}
\boldsymbol{\hat{w}}_{k} = \frac{1/\boldsymbol{\tilde{w}}_{l}}{\sum_{l=1}^{2L}(1/\boldsymbol{\tilde{w}}_{l})}
\end{equation}
The overall procedure of the optimization of the linearly combined base classifiers can be denoted as Algorithm \ref{alg:GradientDescentLinearCombinationClassifier}.\\
\begin{algorithm}[H]
\KwData{Individual Classifiers $\boldsymbol{f}_{l}(\cdot)$, Corresponding Balanced Data set $\boldsymbol{X}$(including $\boldsymbol{\hat{X}}$ and $\boldsymbol{\check{X}}$), where $l=1,2,...,2L$}
\KwResult{Trained Combination of Models $F(\cdot)$}
Initialize the linear combination model with equation \ref{InitializationOfCombinationWeights}\;
\For{$t \gets 1$ \KwTo \textnormal{Max-Step}}{
  Compute the current gradient with equation \ref{equ:GradientOfLinearCombination}\;
  Update the weights with equation \ref{WeightUpdateFormula}\;
  Compute the current cross-entropy loss with equation \ref{equ:CrossEntropyLossLinearCombination}\;
	\If{\text{Cross-Entropy Loss Change} $\leq$ tol}{
    	break\;
	}
}
\caption{Stochastic Gradient Descent Training of Linear Ensemble of Base Classifiers}
\label{alg:GradientDescentLinearCombinationClassifier}
\end{algorithm}
\subsection{Weight Computation and the Probabilistic Prediction}
\label{subsec:WeightInterpolation}
The preceding three sections have discussed generating balanced sub-datasets, and training individual and combined classifiers. As a dynamic ensemble method, the coefficient of each individual classifier should be on-the-fly, according to the test instance(s). Because we have multiple `base' classifiers, and each of them is trained by its corresponding subset, it makes sense to pick the most effective ones, based on the test data. Our selection strategy is to compute the 'distance' (denoted by the likelihood) between the specific test instance and the Gaussian centroids. Intuitively, a higher likelihood indicates a more significant impact on the test instance, and we can assign more `credit' to the corresponding classifiers. Based on the above idea, we propose computing the on-the-fly weights with two types of likelihoods, based on the logarithm and exponential (original) forms, respectively. More concretely, for any test instance $\boldsymbol{x}^{*}$, the component of the normalized likelihood $\boldsymbol{l}^{*}$ will be as follows:\\
\begin{equation}
\label{equ:loglikelihood-explikehood}
\boldsymbol{l}^{*} = 
\begin{cases}
\frac{\log\mathcal{N}(\boldsymbol{x}^{*}|\boldsymbol{\mu}_{l},\boldsymbol{\Sigma}_{l})}{\sum_{l=1}^{2L}\log\mathcal{N}(\boldsymbol{x}^{*}|\boldsymbol{\mu}_{l},\boldsymbol{\Sigma}_{l})}, & \text{ Log-likelihood}\\
\frac{\mathcal{N}(\boldsymbol{x}^{*}|\boldsymbol{\mu}_{l},\boldsymbol{\Sigma}_{l})}{\sum_{l=1}^{2L}\mathcal{N}(\boldsymbol{x}^{*}|\boldsymbol{\mu}_{l},\boldsymbol{\Sigma}_{l})}, & \text{ Exp-likelihood}
\end{cases}
\end{equation}
In the above equation, because the global geometry is considered (especially in the normalization), the on-the-fly weights are regularized. Finally, for the purpose of the SGD-induced regularization, we will interpolate the on-the-fly weights with the learned static ones:
\begin{equation}
\label{equ:WeightInterpolation}
\boldsymbol{w}^{*} = \lambda \boldsymbol{l}^{*} + (1-\lambda) \boldsymbol{w}
\end{equation}
where the “$+$” operation stands for pairwise summation. $\lambda$ is the interpolation parameter, and the optima can be found via the training or the validation data through grid search. It is easy to verify that the results computed by equation \ref{equ:WeightInterpolation} satisfy the condition that the sum should be equal to 1, and the value of each coefficient will be in the interval $[0,1]$. \par
Following the above procedure, the algorithm will compute a probabilistic value $\boldsymbol{w}^{T}f(\boldsymbol{x}^{*}) \in [0,1]$ for each test point $\boldsymbol{x}^{*}$. The output can be regarded as the probability of $p(\boldsymbol{y}^{*}=1|\boldsymbol{x}^{*})$, and instead of simply setting all samples greater than $0.5$ as $1$, and less, as $0$, the threshold can be tuned, based on the following equation:
\begin{equation}
\label{equ:ThresholdDetermination}
F(\boldsymbol{x}^{*}) = 
\begin{cases}
1, \boldsymbol{w}^{T}f(\boldsymbol{x}^{*}) \geq \delta \\
0, \boldsymbol{w}^{T}f(\boldsymbol{x}^{*}) < \delta
\end{cases}
\end{equation}
where $\delta$ can be regarded as a ”threshold value,” and the optima can be found via the validation data through grid search.\par
The overall procedure of the proposed AER with the XGBoost implementation (AER-XGBoost) can be depicted as algorithm \ref{alg:OverallAlgorithm}. \\
\begin{algorithm}[H]
\KwData{Imbalanced Data Set with Data Separated as $\boldsymbol{X}_{n}$ and $\boldsymbol{X}_{k}$; Test Data Set $\boldsymbol{X}^{*}$}
\KwResult{Label Predictions using the overall model $F(\boldsymbol{X}^{*})$}
Performe the GMM-fitting for the majority class, and generate $2L$ balanced subsets, based on the algorithm \ref{alg:MixtureOfGaussianAndSubsampling}\;
\For{$l \gets 1$ \KwTo $2L$}{
	Initialize the current XGBoost Model $\boldsymbol{f}_{l}(\cdot)$\;
	\For{$t \gets 1$ \KwTo \textnormal{Maximum-Boosting-Depth}}{
    	Update the XGBoost Model according to equations \ref{PseudoResidualComputation}-\ref{equ:alphaValueOptimization}\;
    }
    Store the current model $\boldsymbol{f}_{l}(\cdot)$\;
}
Linearly combining the models with coefficients learned from Algorithm \ref{alg:GradientDescentLinearCombinationClassifier}\;
Use the validation or the training data to find the optimal $\lambda$ value and $\delta$ value \par
\For{\textnormal{each test point $\boldsymbol{x}^{*}$}}{
	Compute the interpolated weight $\boldsymbol{l}^{*}$ of instance $\boldsymbol{x}^{*}$ with equation \ref{equ:loglikelihood-explikehood}\;
    Update the weights $\boldsymbol{w}^{*}$ through equation \ref{equ:WeightInterpolation}\;
    Predict the current label $F(\boldsymbol{x}^{*})$ with equation \ref{equ:ThresholdDetermination}\;
}
\caption{ Overall Algorithm}
\label{alg:OverallAlgorithm}
\end{algorithm}

\section{Theoretical Analysis of the AER}
\label{ComplexitySection}
In this section, we will demonstrate that the proposed AER method has advantageous time and memory complexity. Specifically, we will show theoretically that, under certain assumptions and for any classifier implemented with the AER framework, the time complexity will be, asymptotically, at least as good as the original implementation, and the asymptotic memory complexity will always be better than the full-batch implementations.\par
To begin with, let us recap the notations used in the AER model. Recall that $m$ denotes the number of instances, and $n$ represents the number of features. $m_k$ and $m_n$ are used for the minority and majority data, respectively. The skew rate here will be denoted as $R$; it is straightforward to deduce that $m_n = R\cdot m_k$. The number of Gaussian centroids is given as $L$, and, in most cases, $L<R$, as it will otherwise miss the purpose of re-sampling (it is possible to simply train balanced sub-sets, and include all the training set). This also implies that $L<m$ as $m=(R+1)m_{k}\geq(R+1)>L$. The number of iterations of the GMM E-M algorithm will be denoted as $t_1$, and the number of iterations of the SGD algorithm is denoted by $t_2$. The time complexity of any machine learning classifier $T(m,n)$ can be denoted as a polynomial of the numbers of instances and features $T(m,n) = O(m^{a}n^{b})$, where $a$ and $b$ should be positive integers. Similarly, we will denote the memory complexity using $M(m,n) = O(m^{a}n^{b})$. We are preoccupied mostly with the complexity of the training process, as this will usually be the part that consumes most of the time and memory.\par
To derive a boundary that is not dependent on the GMM-fitting or SGD part, and draw fair comparisons between the AER-implemented methods and the original methods, the analysis will be based on the assumption that the GMM covariance inversion and likelihood will be estimated through the diagonal covariance approximation. This will remove high-order terms of $n$, and reduce the time complexity of computing the GMM to $O(mn)$, as the inversion and multiplication of the covariance can be completed within $O(n)$ time. Further, we assume the choice of $\lambda$ and $\delta$ is based on the validation set, and its size is considerably smaller, with the condition $m=\Omega(m_{v}L)$, where $m_{v}$ is the number of validation data points.
\subsection{Time Complexity}
For any machine learning method with polynomial training time complexity $T(m,n) = O(m^{a}n^{b})$, the AER time complexity can be denoted as $T^{AER}(m,n)$. Under the assumptions stated above, the following theorem can be derived:
\begin{theorem}
\label{thm:time}
Assuming $t_1 = O(\log(m))$ and $t_2 \leq n\cdot R/L$, the following property holds: If $a=1$, which means $T(m,n) \in \tilde{O}(mn^{b})$, then, there will be $T^{AER}(m,n) \in \tilde{\Theta}(T(m,n))$; otherwise, if $a \geq 2$, which means $T(m,n) \in \Omega(m^2n^{b})$, then, there will be $T^{AER}(m,n) \in o(T(m,n))$. Both $\tilde{O}(\cdot)$ and $\tilde{\Theta}(\cdot)$ hide polylog terms.
\end{theorem}
\begin{proof}
The theorem can be proved by a simple analysis. The memory complexity of the AER method can be decomposed into four parts: the complexity of computing the GMM model, the training complexity of the individual classifiers, the SGD, and the validation part to obtain the optimal $\delta$ and $\lambda$. Each part will have the following complexity:
\begin{itemize}
\item Fitting the GMM model. The algorithm will fit the $2L$ Gaussian distributions; it will take $L\cdot O(t_{1}\frac{m}{L}n) + L\cdot O(t_{1}\frac{m}{R}n)$ to fit the models under diagonal covariance. The overall complexity will be $O(t_{1}mn)$. Observe that $L<R$ is used in the derivation.
\item Training individual classifiers. For the first type of re-sampled data, the number of training instances will be $\hat{m} = O(\frac{m_n}{L}+m_{k}) = O(\frac{R}{L}m_{k}) = O(\frac{m}{L})$; for the second type of re-sampled data, the amount of training samples will be $\hat{m}=2.5m_{k}= O(m_{k})= O(\frac{m}{R})$. Given the polynomial-form time complexity $T(m,n) = O(m^{a}n^{b})$, the complexity of this part will be $L\cdot[O({(\frac{m}{L})}^{a}n^{b})+O({(\frac{m}{R})}^{a}n^{b})] = O(L{(\frac{m}{L})}^{a}n^{b}) = O({\frac{m^a}{L^{a-1}}}n^{b})$.
\item Stochastic gradient descent. This part will take $O((m_{n}+b_{N})Lt_2)$, where $b_{N}$ is the batch size of the SGD, and $t_2$ is the number of iterations. $b_{N}$ is a constant, and can therefore be hidden asymptotically, resulting in $O(m_{n}Lt_{2}) = O(\frac{mLt_{2}}{R})$ runtime.
\item Validation of the optimal $\lambda$ and $\delta$ parameters. Under the option of diagonal covariance approximation, the likelihood estimation of a single data point will be $O(Ln)$. It is estimated that all the sets are $O(Lm_{v}n)$. The optimal $\lambda$ and $\delta$ values need to be obtained via multiple running times; however, the factor can be hidden, as it will be a constant.
\end{itemize}
Summarizing the above terms, the overall complexity will be $O(t_{1}mn + {\frac{m^a}{L^{a-1}}}n^{b} + \frac{mLt_2}{R} + Lm_{v}n)$. Because the condition is given as $t_2 \leq \frac{nR}{L}$ and $m,n \geq 1$, the third term can be hidden. Furthermore, because we assumed a large training set and a small validation set with $m=\Omega(m_{v}L)$, the final part can be hidden, and the complexity will be $T^{AER}(m,n) = O({\frac{m^a}{L^{a-1}}}n^{b})$. \\
Now for the two cases: 
\begin{itemize}
\item If $a=1$, it could deduced that $T^{AER}(m,n) =O(t_{1}mn+{\frac{m}{L^{0}}}n^{b}) = O(m\log(m)\cdot n+mn^{b}) = \tilde{\Theta}(T(m,n))$. 
\item If $a\geq 2$, there will be $T^{AER}(m,n) = O(m\log(m)\cdot n+{\frac{m^a}{L^{a-1}}}n^{b}) = o(m^{a}n^{b}) = o(T(m,n))$.
\end{itemize}
\end{proof}
Because the E-M algorithm used to compute the GMM model converges super-linearly, the assumption of $t_{1}=O(\log(m))$ is not far from the reality. Theorem \ref{thm:time} indicates that by re-sampling the dataset, the proposed AER method can reduce the time complexity, when the original complexity is super-linear, with respect to the amount of data $m$, and will not be worse than the original full-batch implementation, when the complexity is linear to $m$.\par
Table \ref{tab:timeComplexityComparison} presents the comparison results of the time complexity among common machine learning classifiers implemented using the original full-batch scheme and the AER framework. From the table, it can be observed that the higher the order of $m$ in an algorithm, the more advantages the AER framework will bring. The GBM, which is the method-of-choice in our base classifier, is also listed in the table, and $|T|$ denotes the number of trees in the algorithm. It can be observed that our implementation of the GBM is based on the XGBoost, a parallelized GBM method; furthermore, our implementation does not conform to the polynomial-time regime of our analysis. Nevertheless, the rigorous analysis of the time complexity provides a convincing proof of the advantage of the proposed AER method.
\begin{table}
\centering
\captionsetup{justification=centering}
\caption{\label{tab:timeComplexityComparison}Comparison of training time complexity between common Machine Learning classifiers implemented via full-batch original scheme and AER}
\begin{tabular}{c|c|c}
\hline
 & original scheme & AER \\ \hline
Naive Bayes & $mn$ & $mn$ \\[1.5ex] 
Decision Tree & $m^{2}n$ & $\frac{m^{2}}{L}n$\\[1.5ex] 
Non-linear SVM & $m^{3}n$ \textbf{or} $m^{2}n$ & $\frac{m^{3}}{L^{2}} n$ \textbf{or} $\frac{m^{2}}{L} n$\\[1.5ex] 
Gradient Boosting & $mn|T|$ & $mn|T|$\\ \hline
\end{tabular}
\end{table}
\subsection{Memory Complexity}
For any machine learning method with polynomial-training memory complexity $M(m,n) = O(m^{a}n^{b})$, the AER memory complexity can be denoted as $M^{AER}(m,n)$. Based on the assumptions stated above, the following theorem can be obtained:\\
\begin{theorem}
\label{thm:memory}
For any $a,b \in N^{+}$, $M^{AER}(m,n) \in o(M(m,n))$
\end{theorem}
\begin{proof}
Similar to the analysis of the time complexity, the memory complexity of the AER is decomposed into four parts:\\
\begin{itemize}
\item Fitting the GMM model. The model needs to store $n$ values under the setting of diagonal variance; thus, the memory complexity will be $O(nL)$.
\item Training the individual classifiers. Similar to the time complexity proof, the two types of subsets will have the number of samples in $O(\frac{m}{L})$ and $O(\frac{m}{R})$, respectively. One difference here is that for the memory complexity, one could use the same memory for every Gaussian component. Thus, the memory complexity will be $O((\frac{m}{L})^{a}n^{b})$.
\item Stochastic gradient descent. One only needs to keep $2L$ slots in the memory, to update the weights such that the memory complexity will be $O(L)$.
\item Validation of the optimal $\lambda$ and $\delta$ parameter. For each Gaussian component, the validation process will take $O(n)$ memory, and each data point will need $O(nk)$. The likelihood of the $m$ data will be stored, which means there should be an additional $O(m)$ complexity. The overall complexity of this part will be $O(m + nk)$.
\end{itemize}
The final complexity will be given as $O(nL+(\frac{m}{L})^{a}n^{b} + L + nL) = O(nL+(\frac{m}{L})^{a}n^{b})$. Further, because $L<m$, the complexity can be simplified as $O((\frac{m}{L})^{a}n^{b})$, with a simple derivation, one could obtain $M^{AER}(m,n) = O((\frac{m}{L})^{a}n^{b}) = O(\frac{m^a}{L^a}n^{b}) = o(m^{a}n^{b}) = o(M(m,n))$
\end{proof}
It can be observed that the theory of memory complexity is a stronger conclusion than that of time. Firstly, it removes the restrictions on the iteration times, and the memory complexity is unconditionally bounded. Secondly, the theorem \ref{thm:memory} proves a strict upper bound with the little-oh notation (which indicate strictly slower growth asymptotically), regardless of the choice of $a$.

\section{Experimental Analysis: the Framework}
\label{ExperimentalFramework}
In this section, we introduce the framework of our empirical analysis for the AER model. We introduce the datasets in section \ref{subsec:data} with their backgrounds and characteristics. The methods compared against the AER model are discussed in section \ref{subsec:method}, and the metrics to evaluate the results are presented in section \ref{subsec:performancemetric}. Finally, we discuss our approaches for statistical testing to validate the significance of the results in section \ref{subsec:statistictestmetric}.

\subsection{Datasets}
\label{subsec:data}
We introduce 12 imbalanced datasets to evaluate the performance of the proposed AER method. Seven of them are from the UCI machine learning repository\footnote{Available Publicly, url: \url{https://archive.ics.uci.edu/ml/index.php}}(Bioassay, Abalone (originally from \cite{alcala2011keel}), Ecoli, US Crime, Wine Quality, Scene and Car Eval), and are regarded as benchmark datasets for imbalanced binary classification. For the Bioassay dataset, we adopt the AID 362 collection, and split it with the train/test ratio of $4:1$. Table \ref{tab:uci-data-stat} shows the statistics of these datasets. In addition, we provide five artificially generated imbalanced datasets sampled from 8,9,10,11,12-center Gaussian Mixture Models\footnote{Available Publicly, url: \url{https://github.com/jhwjhw0123/GMM-Generated-data-imbalance-classification}}, respectively. Because of space limitation, we select the Bioassay, Abalone and the $9$-center GMM-based dataset to demonstrate the detailed information, including the evaluation metrics, the hypothetical testing results, and the curves of the changes on accuracy with different threshold parameter $\delta$. For the Bioassay dataset, we further plot the SGD-trained weights and the change in accuracy, with respect to the interpolation parameter $\lambda$, to offer more insights into the model. For the remaining five UCI datasets and other variations of the GMM-generated data, we only report the results of the evaluation metrics and the results of the statistical tests. \par

\begin{table}[]
    \centering
    \caption{Statistics of seven Datasets from UCI Machine Learning Repository}
    \begin{tabular}{c|c|c|c}\hline
    
        Dataset Name & Number of Samples & Imbalance Ratio & Number of Features \\\hline
        Bioassay & 4279 & 1:70 & 118 \\
        Abalone & 4174 & 1:129 & 8\\
        Ecoli & 336 & 1:8.6 & 7\\
        US Crime & 1994 & 1:12 & 100 \\
        Wine Quality & 4898 & 1:26 & 11\\
        Scene & 2407 & 1:13 & 294\\
        Car Eval & 1728 & 1:12 & 21\\\hline
    \end{tabular}

    \label{tab:uci-data-stat}
\end{table}

Table \ref{tab:uci-data-stat} shows the meta-information about the UCI datasets we used in this section. We observe that among the datasets, Bioassay and Abalone have the most significant imbalance ratio. Thus, presenting more details through experiments with these two datasets can help us understand the merits of the AER more clearly. Among all the other datasets, Ecoli is significant for its small size, which can lead to problems for some data-hungry machine learning methods (e.g. focal loss neural Networks). The Scene dataset is of a higher dimension (a greater number of features), and in the corresponding section, we will demonstrate that the AER can perform particularly well, when such rich-information data is provided. \par

The GMM-based data are introduced to verify that the AER model follows the intuition, because the data geometry is truly from a GMM distribution. A set of 8000 samples is generated through the sk-learn \texttt{Make-Classification()} method, and the imbalance ratio is 1:79, which indicates that 7900 samples are labeled as “0,” and 100, as “1.” Notice that the “number of Gaussian centers” is applied to both the majority and minority data, which means the 100 positive-labeled samples are also from the GMM with 8-12 clusters (depending on the variation). This poses a significant challenge; the algorithms have to learn a complex decision boundary, while preserving a qualified generalization ability. As shall be seen in the corresponding section, the proposed method performs well on this dataset, whereas some other methods, including some state-of-the-art models, completely lose their ability to grasp anything meaningful.

\subsection{Compared Methods}
\label{subsec:method}
We compare seven methods with the proposed AER model with the XGBoost implementation (AER-XGBoost). Among the seven, three are benchmark binary classification methods, which are the (cost-sensitive) SVM, (cost-sensitive) decision tree and a (3-layer) focal loss neural network. The neural network model is engineered with the recently proposed focal loss (\cite{lin2018FocalLoss}) to tackle the label-imbalance problem. To introduce some more novel and advanced methods for comparison, we adopted three more recent methods, the XGBoost (\cite{Chen2016XGBoost}), LightGBM (\cite{ke2017lightgbm}), and Focal-XGBoost (\cite{wang2019imbalance}). Among these three methods, the XGBoost is also adopted as the sub-classifier of the AER model; we can therefore straightforwardly evaluate the benefits brought by the AER framework (in comparison with the vanilla version and other imbalance-augmentation methods, e.g., focal loss). Finally, we also report the results based on the pure dynamic ensemble, which can help validate the contribution of the regularization introduced by the AER. \par

It is noticeable that apart from the UCI Bioassay data, all the other datasets are not partitioned into training/testing subsets. Thus, in the experiments, we split the training, validation, and testing data in the ratio of $3:1:1$. For the Bioassay dataset, part of the results (cost-sensitive SVM and cost-sensitive decision tree) are retrieved from \cite{schierz2009BioassayData}. It is necessary to tune their parameters to the optimal; however, we \emph{cannot} reproduce their results, because we do not know these parameters. Therefore, for the purpose of fair comparison, we directly reported their results on the cost-sensitive SVM and cost-sensitive decision tree algorithms. \par

The parameters of both the AER model and the methods of comparisons are tuned to the optimal as we know from a fixed range (except the number of Gaussian distributions for the Dynamic Ensemble and the AER models, since this parameter is manifold-specific and should be searched from a dynamic range). Notice that since the distributions across datasets are different, the final parameters to be determined vary among experiments. The details of the final parameters in each experiment can be found among the results in section \ref{ExperimentSection}. For the convenience of the readers, we also include the range of parameters for all the methods in the appendix.

\subsection{Performance Metrics}
\label{subsec:performancemetric}
For an ordinary classification problem, the accuracy can simply be used as the sole metric to evaluate performance. However, for label-skewed data, the algorithm often achieves a satisfying accuracy, even by simply predicting every instance as the majority class. Thus, in this scenario, the spotting results of the majority and the minority data should be examined, respectively. Specifically, if one regards the minority data as Positive (P), and majority as Negative (N), then, combining the prediction results and ground-truth labels will yield four prediction outcomes: true positive (TP), false positive (FP), true negative (TN), and false negative (FN). As a conventional analytical approach, precision and recall will be introduced to evaluate the quality of the classification of the majority/minority data. The computation of precision and recall metrics are given as follows:\\
\begin{equation}
\label{PrecisionRecallFormula}
\begin{aligned}
&\text{precision} &= 
\begin{cases}
\frac{TP}{TP+FP}, \text{ Minority Data}\\
\frac{TN}{TN+FN}, \text{ Majority Data}
\end{cases}\\
&\text{recall} &= 
\begin{cases}
\frac{TP}{TP+FN}, \text{ Minority Data}\\
\frac{TN}{TN+FP}, \text{ Majority Data}
\end{cases}\\
\end{aligned}
\end{equation}
Notice that, in this study, the concept of 'precision' and 'recall' are extended to class-specified metrics, in contrast to focusing only on the positive(minority) samples, as in the conventional statistical analysis. Thus, in our experiments, both the majority and minority recalls are reported. On the premise of sufficient recall, the TP-FP ratio can also be employed to evaluate the quality of the label-skewed data classification:
\begin{equation}
\label{TPFPratio}
\begin{split}
\text{TP-FP Ratio} &= \frac{TP/(TP+FN)}{FP/(TN+FP)}\\
&= \frac{\text{recall}_{k}}{1-\text{recall}_{n}}
\end{split}
\end{equation}
where $\text{recall}_{k}$ and $\text{recall}_{n}$ stand for the recall of the minority and majority classes, respectively. To evaluate the overall quantities of precision and recall, the $F_{1}$ score and G-Mean are introduced. The $F_{1}$ score and G-Mean are computed as follows:\\
\begin{equation}
\begin{aligned}
&F_{1} = 2\frac{\text{precision}*\text{recall}}{\text{precision}+\text{recall}}\\
&\text{G-Mean} = \sqrt{\text{precision}*\text{recall}}
\end{aligned}
\end{equation}
The $F_{1}$ score and G-Mean are commonly used metrics in imbalanced classification problems. The G-mean is usually a visibly more consistent metric, and could therefore provide more reliable information in our experiments \cite{espindola2005extending,luo2018feature}.\par
In addition, the “balanced accuracy” is also introduced as follows:\\
\begin{equation}
\label{BalancedAccuracy}
\text{Balanced Accuracy} = \frac{\text{recall}_{k}+\text{recall}_{n}}{2}
\end{equation}
\par
In summary, mainly, the following metrics will be used in this study: the recall of both the majority and minority classes, the TP-FP ratio, $F_{1}$ score, and G-mean of the minority class, and the TP-FP ratio, and the Balanced Accuracy as the overall performance evaluation.

\subsection{Statistical Testing}
\label{subsec:statistictestmetric}
To further validate the performance superiority of the AER-XGBoost method, the McNemar's test  and Wilcoxon signed-rank test are applied to all the datasets, except Bioassay. The UCI Bioassay dataset is left out, because the metrics are directly retrieved from the literature, and we cannot reproduce their predictions as we do not know their data split and parameter setups. Both tests are conducted on AERs with log- and exp-likelihoods to verify the statistical significance of their performance supremacy over other methods, including the 
cost-sensitive SVM, cost-sensitive decision tree, focal loss neural network, LightGBM, and the plain and focal XGBoosts.

The McNemar's test and Wilcoxon signed-rank test are non-parametric methods commonly used in binary classification problems \cite{pal2013kernel,wilcoxon1992individual}. The idea behind McNemar's test  is based on verifying 'if the two approaches make mistakes on the same part of the sample'. The effectiveness of the test in binary classification tasks was comprehensively discussed in \cite{dietterich1998approximate}. The Wilcoxon signed-rank test is motivated by Student's T-test for paired samples, and it is utilized to verify if the difference between the pairs follows a symmetric distribution around zero. The ranked nature of Wilcoxon test makes it ideal for verifying classification results \cite{graczyk2010nonparametric}. The tests are implemented based on the \textit{Statsmodels} package in Python \cite{seabold2010statsmodels}. For each test, the $\chi^{2}$ statistics and the $p$-values are reported together with possibility of successfully rejecting the null hypothesis.

\section{Experimental Analysis: the Results and Discussions}
\label{ExperimentSection}

In this section, we present and analyze the experimental results of the proposed AER method. As introduced in section \ref{subsec:data} and section \ref{subsec:method}, seven compared methods are implemented on twelve imbalanced datasets. Limited by space, UCI Bioassay and Abalone 19 datasets are selected for primary demonstration, including the performance evaluation of AER with respect to the change of related parameters, also a comprehensive table illustrating performance comparison between AER and other methods. For GMM-generated data with various informative feature number and the rest five UCI imbalanced dataset, we provide the results with compared methods. Two statistical tests: McNemar's test and Wilcoxon test are applied to all datasets.

\subsection{UCI Bioassay}
In the experiment, the number of Gaussian centroids for the AER is chosen from a set of $\{7,8,9\}$. After performing validation with the minimum BIC value based on equation \ref{equ:BIC_computation_MoG}, the final number of Gaussian centroids is optimized as $8$. This leads to the total number of base classifiers being 16, of which the first eight are trained on the majority-dominating subsets; the rest of them are fitted with the nearly balanced subsets.
\begin{figure}
\centering
\includegraphics[width=1.0\textwidth]{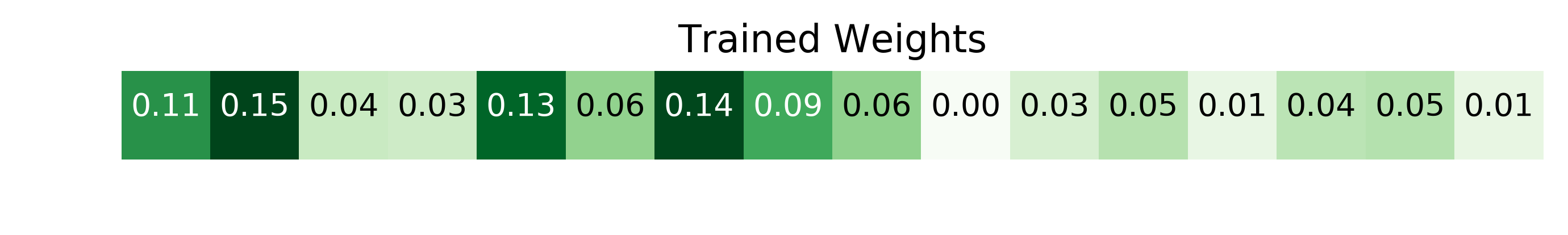}
\caption{\label{fig:TrainedWeights}The trained weights of the algorithm for UCI Bioassay data (round to 2 decimals). The first 8 weights corresponding to XGBoost classifiers are trained on majority-dominated subset, while the later 8 are fitted on nearly label-balanced subsets.}
\end{figure}
\par
The distribution of the trained weights is plotted in Figure \ref{fig:TrainedWeights} (rounded to two decimals places, for the convenience of plotting). From the figure, we can see that the classifiers trained on the majority-dominated data generally have larger weights (with larger values and darker colors) because these classifiers better represent the global geometry. Nevertheless, the weights from the balanced dataset also have indisputable influences on the overall prediction. 
\par
\begin{figure}
\centering
\includegraphics[width=1.0\textwidth]{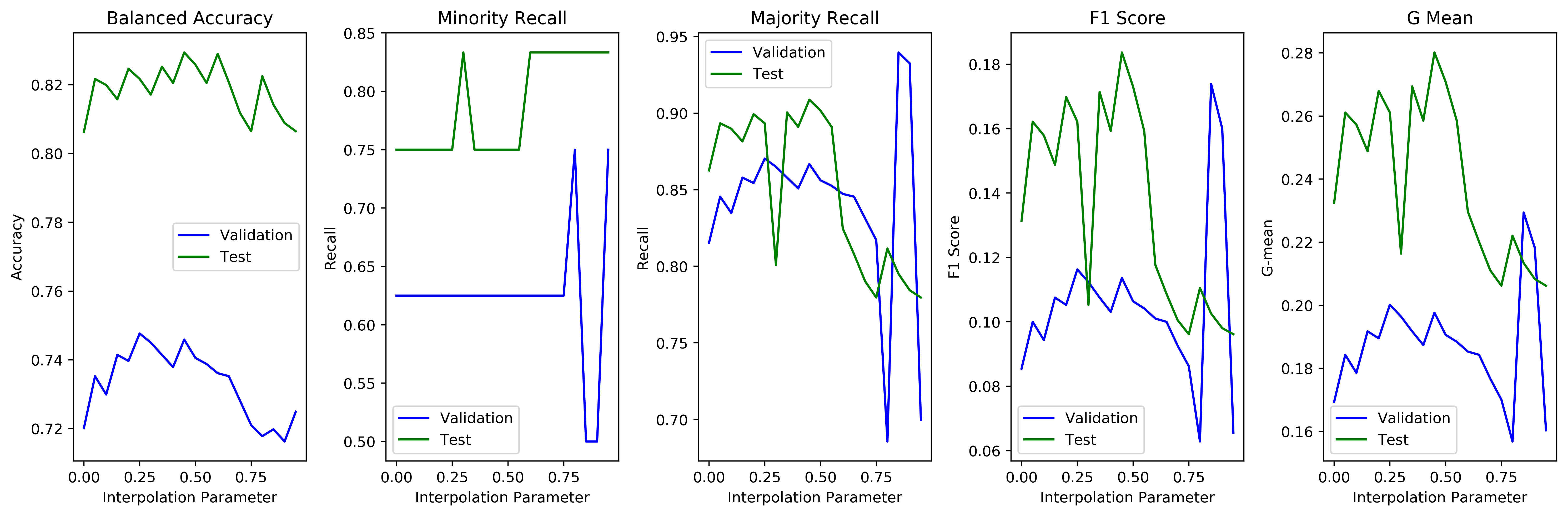}
\caption{\label{fig:WeightInterpolationLog}The performance of the overall algorithm with respect to the change of the interpolation parameter $\lambda$ in the equation \ref{equ:WeightInterpolation} on the \textbf{UCI Bioassay dataset}. Log likelihood is adopted to compute the likelihood of each test data.}
\end{figure}
\begin{figure}
\centering
\includegraphics[width=1.0\textwidth]{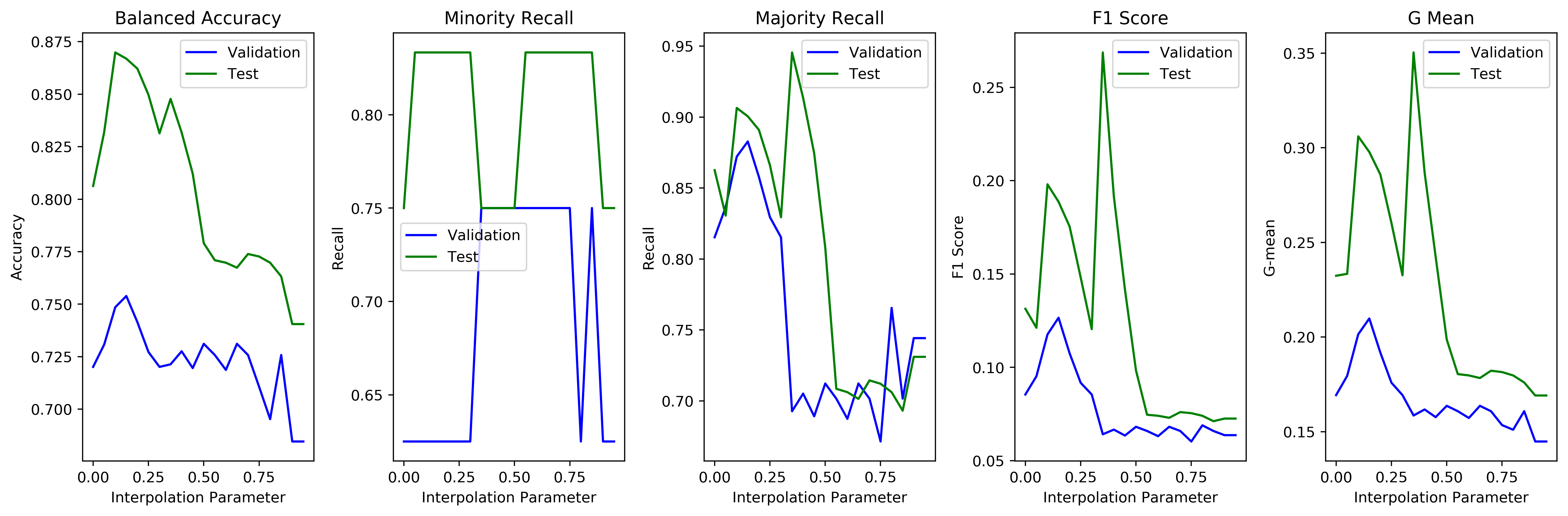}
\caption{\label{fig:WeightInterpolationExp}The performances of the overall algorithm with respect to the change of the interpolation parameter $\lambda$ in Equation \ref{equ:WeightInterpolation} on the \textbf{UCI Bioassay dataset}. Exponential likelihood is adopted to compute the likelihood of each test data.}
\end{figure}

The grid search results of the different values of the interpolation parameter $\lambda$ are shown in Figures \ref{fig:WeightInterpolationLog} and \ref{fig:WeightInterpolationExp}. The unit measurement of the $\lambda$ is $0.05$, and $\lambda=1$ indicates that the system only relies on the learned weight vector; which $\lambda=0$ indicates that the model solely depends on trained coefficients. From the curve, it can be found that neither of the above setups will yield results as good as the interpolated model, which further demonstrates the effectiveness of the AER framework. In addition, Table \ref{tab:interpolationOptimal} shows that the optimal $\lambda$ obtained by the validation data is close to the `real' optimal $\lambda$, based on the test data.
\begin{table}
\centering
\captionsetup{justification=centering}
\caption{\label{tab:interpolationOptimal}The comparison between validation-based and true testing optimal $\lambda$ on the \textbf{UCI Bioassay dataset}.}
\begin{tabular}{|l|p{3cm}|p{3cm}|p{3cm}|}
\hline
 & Validation optimal $\lambda$ & Corresponding test balanced-accuracy & Optimal test balanced-accuracy \\ \hline
Log-likelihood & 0.25 & 0.8217 & 0.8294 \\ \hline
Exp-likelihood & 0.15 & 0.8669 & 0.8698 \\ \hline
\end{tabular}
\end{table}
\par
\begin{figure}
\centering
\includegraphics[width=1.0\textwidth]{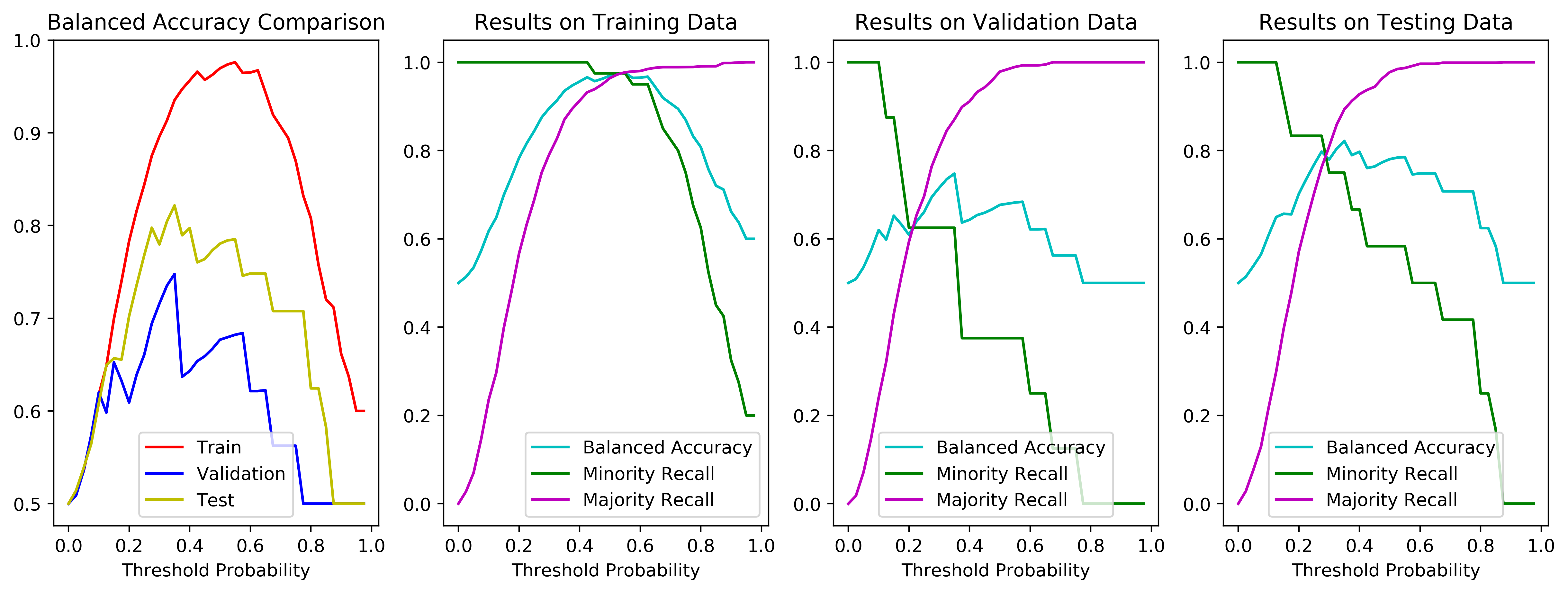}
\caption{\label{fig:LogLikelihoodPerformance}The performances of the proposed dynamic ensemble algorithm with respect to the change of the threshold parameter $\delta$ in Equation \ref{equ:ThresholdDetermination} on the \textbf{UCI Bioassay dataset}. Log likelihood is adopted to compute the likelihood of each test data.}
\end{figure}
\begin{figure}
\centering
\includegraphics[width=1.0\textwidth]{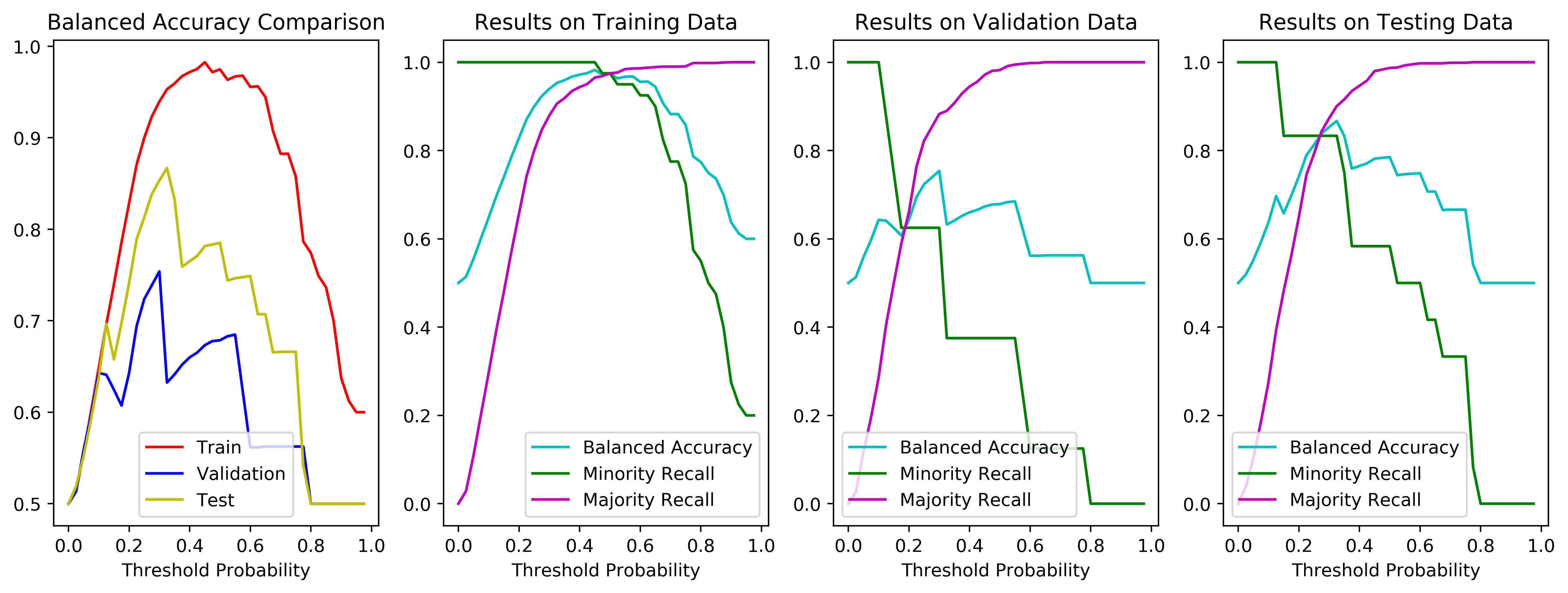}
\caption{\label{fig:ExpLikelihoodPerformance}The performances of the proposed dynamic ensemble algorithm with respect to the change of the threshold parameter $\delta$ in Equation \ref{equ:ThresholdDetermination} on the \textbf{UCI Bioassay dataset}. Exponential likelihood is adopted to compute the likelihood of each test data.}
\end{figure}

After obtaining the optimal values of $\lambda$, one can examine the optimal value of $\delta$ in Equation \ref{equ:ThresholdDetermination}, based on the training or validation data. With the selection metric stated above, for this dataset, the difference between the average log-likelihoods of the validation and testing data is $1602.56$, and the same metric between the training and testing data is $8663.5$. Therefore, the validation data is selected to determine $\delta$. With the interpolation parameter in Table \ref{tab:interpolationOptimal}, the performance, with respect to the changing value of $\delta$ is illustrated in Figures \ref{fig:LogLikelihoodPerformance} and \ref{fig:ExpLikelihoodPerformance}.
\par
\begin{figure}[!h]
\centering
\includegraphics[width=0.7\textwidth]{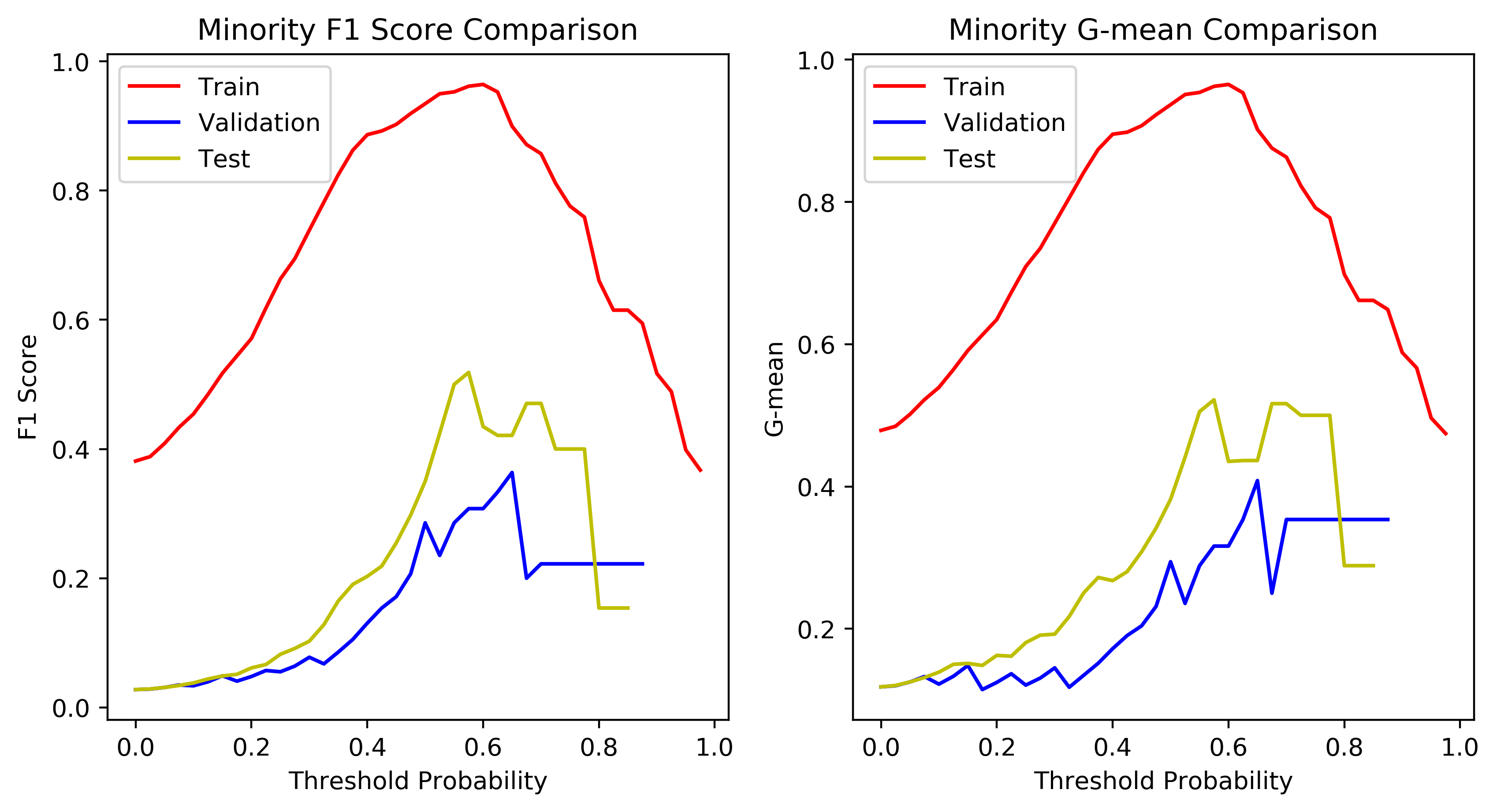}
\caption{\label{fig:LogLikelihoodF1Gmean}Change in $F_{1}$ score and G-mean of the minority data with respect to the changing value of $\delta$ in Equation \ref{equ:ThresholdDetermination} on the \textbf{UCI Bioassay dataset}. Log likelihood is adopted to compute the likelihood of each test data.}
\end{figure}
\begin{figure}[!h]
\centering
\includegraphics[width=0.7\textwidth]{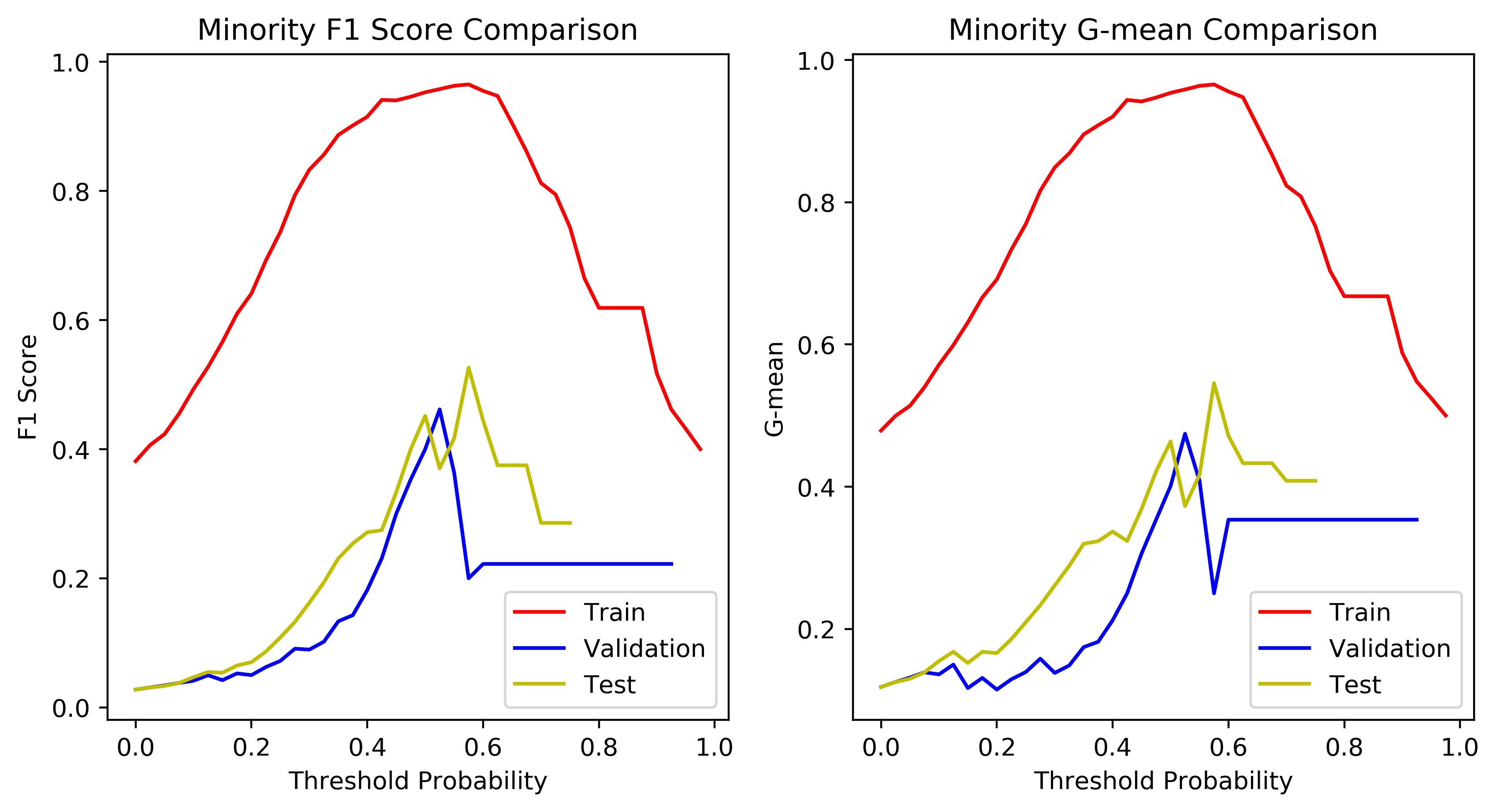}
\caption{\label{fig:ExpLikelihoodF1Gmean}Change in $F_{1}$ score and G-mean of the minority data with respect to the changing value of $\delta$ in Equation \ref{equ:ThresholdDetermination} on the \textbf{UCI Bioassay dataset}. Exponential likelihood is adopted to compute the likelihood of each test data.}
\end{figure}

The optimal $\delta$ value, based on the validation data, is $0.35$ for the log-likelihood AER, and $0.30$ for the exp-likelihood AER. The selection of the parameter shows that the overall algorithm favors spotting the majority instances over minority samples, as the optimal $\delta$ values under both settings are less than $0.5$. However, given the condition that the minority instances are sparse in the validation and test sets, the results are satisfactory. To provide some further insights, the statistics of the F1 score and G-Mean are given in Figures \ref{fig:LogLikelihoodF1Gmean} and \ref{fig:ExpLikelihoodF1Gmean}. The figures denote the change in the F1 score and G-mean metrics for different $\delta$ values with the log- and exp- likelihoods, respectively.

\begin{figure}[!hb]
\centering
\includegraphics[width=1.0\textwidth]{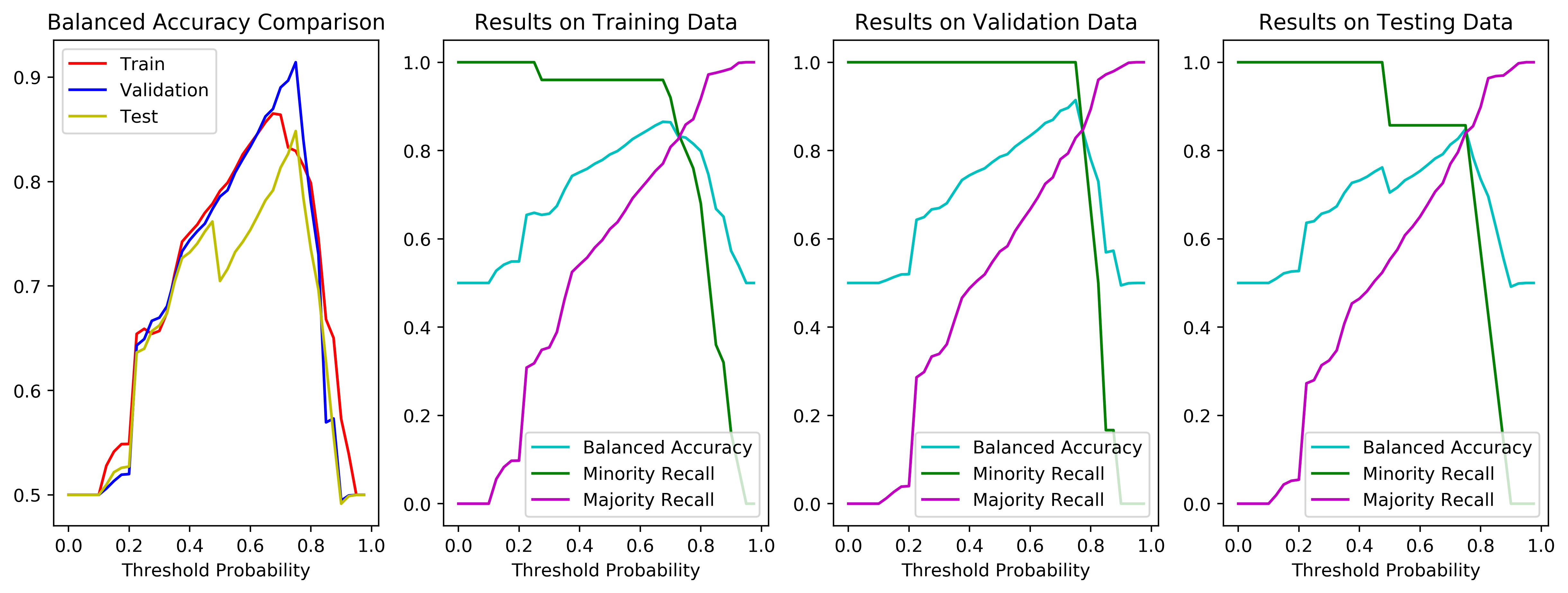}
\caption{\label{fig:AbaloneLogLikelihoodPerformance}Performance of the proposed dynamic ensemble algorithm with respect to changes in the threshold parameter $\delta$ in Equation \ref{equ:ThresholdDetermination} on the \textbf{Abalone 19 dataset}. Log likelihood is adopted to compute the likelihood of each test data.}
\end{figure}
\begin{figure}[!hb]
\centering
\includegraphics[width=1.0\textwidth]{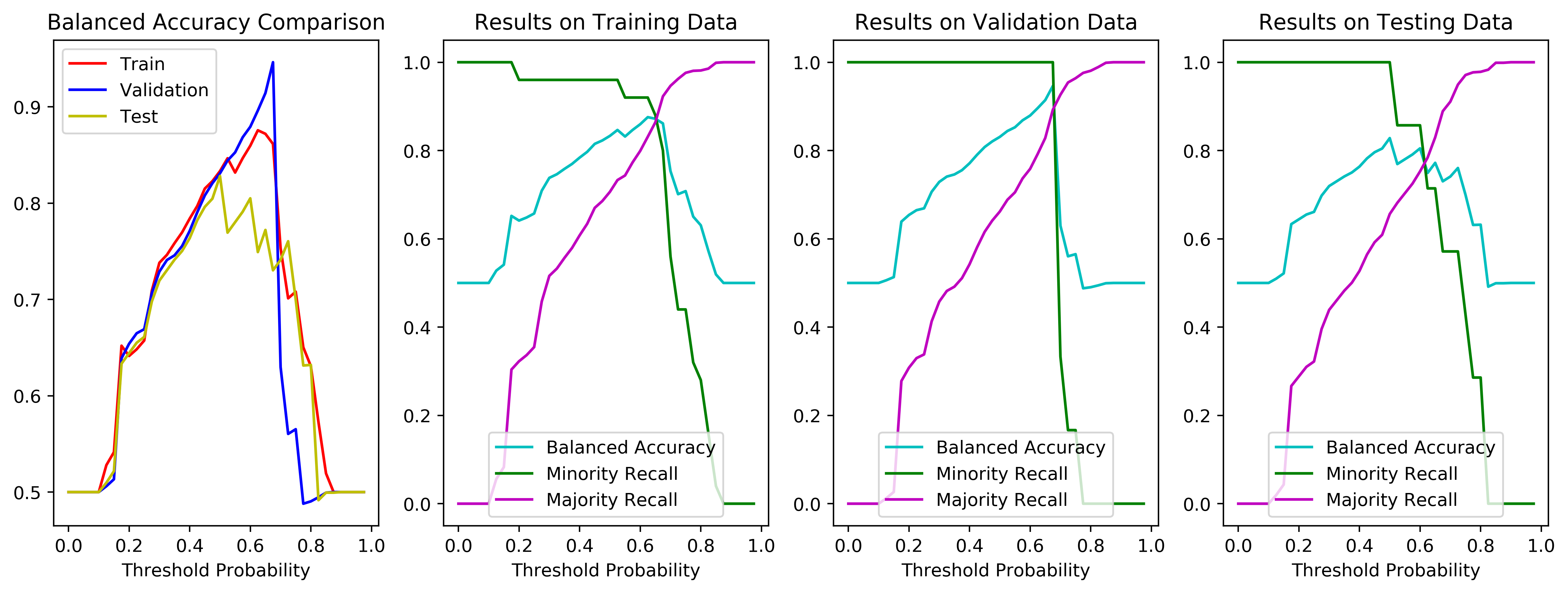}
\caption{\label{fig:AbaloneExpLikelihoodPerformance}Performances of the proposed dynamic ensemble algorithm with respect to changes in the threshold parameter $\delta$ in Equation \ref{equ:ThresholdDetermination} on the \textbf{Abalone 19 dataset}. Exponential likelihood is adopted to compute the likelihood of each test data.}
\end{figure}

\begin{figure}[!hb]
\centering
\includegraphics[width=1.0\textwidth]{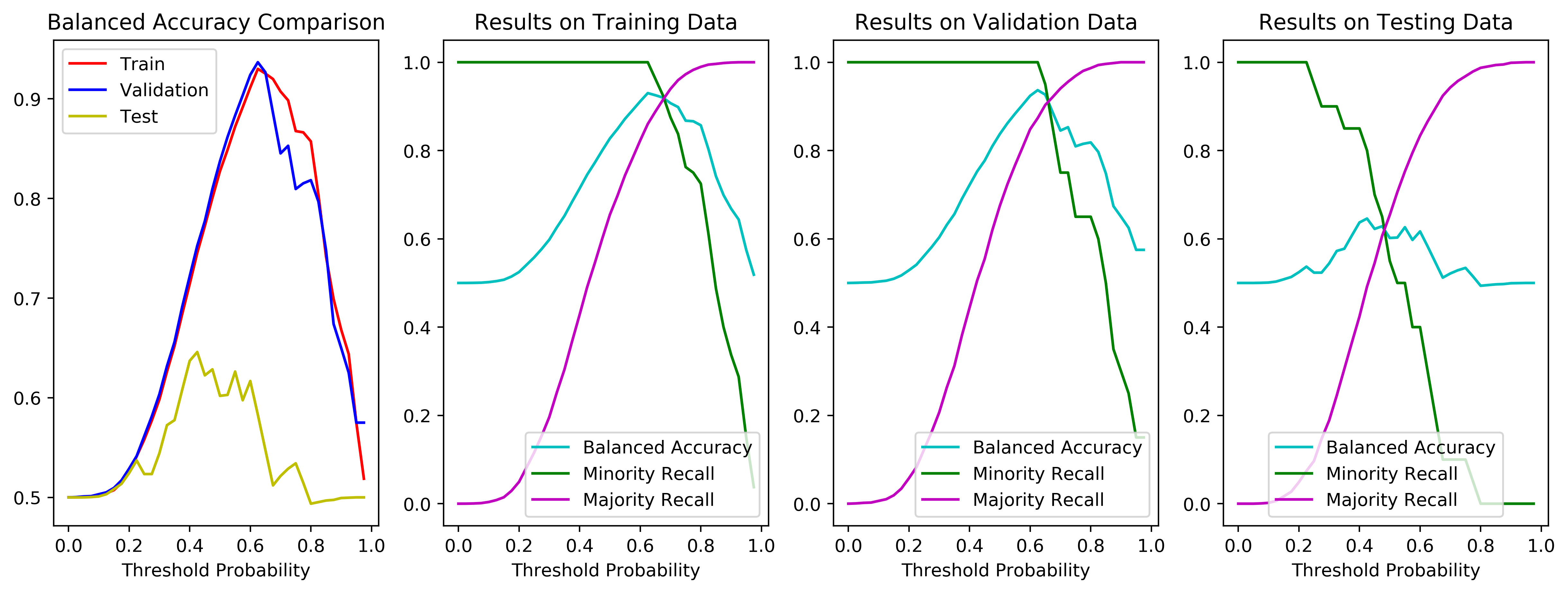}
\caption{\label{fig:GMMLogLikelihoodPerformance}Performance of the proposed dynamic ensemble algorithm with respect to changes in the threshold parameter $\delta$ in Equation \ref{equ:ThresholdDetermination} on the \textbf{GMM-generated dataset}. Log likelihood is adopted to compute the likelihood of each test data.}
\end{figure}
\begin{figure}[!ht]
\centering
\includegraphics[width=1.0\textwidth]{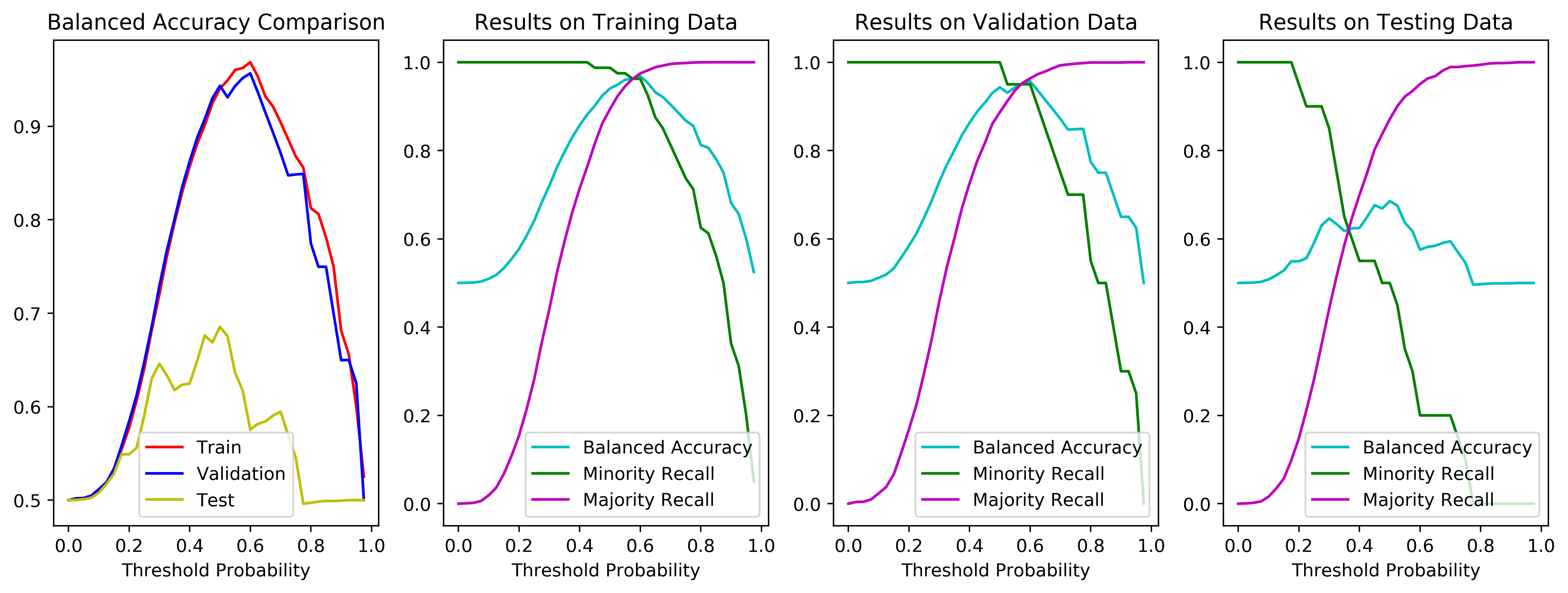}
\caption{\label{fig:GMMExpLikelihoodPerformance}Performances of the proposed dynamic ensemble algorithm with respect to changes in the threshold parameter $\delta$ in Equation \ref{equ:ThresholdDetermination} on the \textbf{GMM-generated dataset}. Exponential likelihood is adopted to compute the likelihood of each test data.}
\end{figure}
\par

Finally, Table \ref{tab:PerformanceComparisonTable} illustrates the performances of the comparison methods mentioned earlier and the AER-XGBoost method with both log- and exp-likelihoods. From Table \ref{tab:PerformanceComparisonTable}, it can be observed that the AER-XGBoost algorithm yields a superior performance, based on multiple metrics. The exp-likelihood-based adaptive ensemble method can yield the best balanced accuracy, and its log-likelihood counterpart yields a better TP-FP ratio, because of a stronger capability in spotting majority instances. Among the compared methods, it can be observed that the focal loss neural network and LightGBM models cannot learn anything meaningful, and a plausible reason for this is that these models tend to overfit, especially when the dataset is small. The XGBoost and focal-XGBoost \cite{wang2019imbalance} demonstrate the strongest performance among the non-AER models. Nevertheless, we find that in the plot of the change in performance, with respect to $\delta$, the range of $\delta$ leading to a competent performance is quite restricted, which means its performance declines drastically, as $\delta$ is slightly larger/smaller than the desired range. This problem is much less significant in the AER models, as we can observe from Figures \ref{fig:LogLikelihoodPerformance} and \ref{fig:ExpLikelihoodPerformance}.
\begin{table}[h!]
\centering
\captionsetup{justification=centering}
\caption{\label{tab:PerformanceComparisonTable}Comparison between the performance of different algorithms on the \textbf{UCI Bioassay dataset}.}
\begin{tabular}{|l|p{1.5cm}|p{1.5cm}|p{1.2cm}|p{1.0cm}|p{1.0cm}|p{1.5cm}|}
\hline
\abcdresults{ }{Balanced Accuracy}{Minority Recall}{Majority Recall}{F1 score}{G-mean}{TP-FP Ratio} \hline
\abcdresults{Cost-sensitive SVM}{79.97\%}{75.00\%}{84.93\%}{ 0.1224}{0.2236}{5.0238} \hline
\abcdresults{Cost-sensitive Decision Tree}{80.08\%}{ 75.00\%}{85.16\%}{0.1241}{0.2253}{5.1048}\hline

\abcdresults{Focal loss neural network}{50.00\%}{0}{100\%}{--}{--}{--} \hline
\abcdresults{LightGBM}{50.00\%}{0}{100\%}{--}{--}{--} \hline
\abcdresults{Plain XGBoost}{41.90\% }{83.33\%}{0.47\%}{ 0.0232}{0.0990}{0.8373} \hline
\abcdresults{Focal-loss XGBoost}{80.77\%}{83.33\%}{78.19\%}{ 0.0971}{0.2073}{3.8225} \hline
\abcdresults{Dynamic Ensemble (log)}{41.90\% }{83.33\%}{0.47\%}{ 0.0232}{0.0990}{0.8373} \hline
\abcdresults{Dynamic Ensemble (exp)}{80.77\%}{83.33\%}{78.19\%}{ 0.0971}{0.2073}{3.8225} \hline
\abcdresults{AER-XGBoost (Log)}{82.16\%}{75.00\%}{89.34\%}{ 0.1622}{0.2611}{7.0333} \hline
\abcdresults{AER-XGBoost (Exp)}{\textbf{85.33\%}}{83.33\%}{87.33\%}{ 0.1550}{0.2668}{6.5732} \hline
\end{tabular}
\end{table}
\subsection{Abalone 19 Data}
The candidate list of the number of Gaussian distributions for the Abalone 19 Dataset is set to $\{11,12,13\}$, and the number $12$ is retrieved through the BIC criteria. The optimal $\lambda$ is determined as $0.15$ for the log-likelihood, and $0.20$ for the exp-likelihood. Further, we again select the $\delta$ values that guarantee the optimal performance on the validation set as the “determined threshold.” The $\delta$ value AER models on the Abalone data is $0.75$ for the log-likelihood, and $0.675$ for the exp-likelihood. The performance, with respect to the changes in the $\delta$ values, is given in Figures \ref{fig:AbaloneLogLikelihoodPerformance} and \ref{fig:AbaloneExpLikelihoodPerformance}. \par
All the methods mentioned in the previous experiment are implemented under the context of the Abalone 19 dataset. Because we no longer retrieve the performance from other literature, the decision tree model is implemented with the SK-learn CART decision tree (almost identical to C4.5 for classification problems), and the SVM model is fine-tuned using the best kernel among \textit{linear}, \textit{RBF} and \textit{Polynomial}. The $\gamma$ parameter for the focal-XGBoost and the focal loss neural network are obtained via a validation grid search, and the final values are set to $2.5$ and $3.0$. \par
\begin{table}
\centering
\captionsetup{justification=centering}
\caption{\label{tab:PerformanceComparisonAbalone}The comparison between the performance of different algorithms on \textbf{Abalone 19 dataset}.}
\begin{tabular}{|p{4.5cm}|p{1.5cm}|p{1.5cm}|p{1.2cm}|p{1.0cm}|p{1.0cm}|p{1.5cm}|}
\hline
\abcdresults{ }{Balanced Accuracy}{Minority Recall}{Majority Recall}{F1 score}{G-mean}{TP-FP Ratio} \hline
\abcdresults{Cost-sensitive SVM}{58.62\%}{28.57\%}{88.66\%}{ 0.0388}{0.0772}{2.5198} \hline
\abcdresults{Cost-sensitive Decision Tree }{56.96\%}{14.29\%}{99.64\%}{0.1818}{0.1889}{--} \hline
\abcdresults{Focal loss neural network}{50.00\%}{0}{100\%}{--}{--}{--} \hline
\abcdresults{LightGBM}{50.00\%}{0}{100\%}{--}{--}{--} \hline
\abcdresults{Plain XGBoost }{50.06\%}{100\%}{0.12\%}{0.0166}{0.0916}{1.0012} \hline
\abcdresults{Focal-loss XGBoost}{ 64.19\%}{42.86\%}{85.52\%}{0.0462}{0.1022}{2.9607} \hline
\abcdresults{Dynamic Ensemble (log)}{71.88\% }{57.14\%}{86.61\%}{ 0.0656}{0.1410}{4.2677} \hline
\abcdresults{Dynamic Ensemble (exp)}{71.88\% }{57.14\%}{86.61\%}{ 0.0656}{0.1410}{4.2677} \hline
\abcdresults{AER-XGBoost (Log)}{\textbf{84.83\%}}{85.71\%}{83.96\%}{0.082}{0.1924}{5.3426} \hline
\abcdresults{AER-XGBoost (Exp)}{73.02\%}{57.14\%}{88.90\%}{0.0777}{0.1543}{5.1491} \hline
\end{tabular}
\end{table}
Table \ref{tab:PerformanceComparisonAbalone} reports the performances of the AER and the compared methods. From the table, it can be observed that the AER method (AER-XGBoost with both log- and exp-likelihoods) outperforms existing algorithms, in terms of the balanced accuracy and G-mean score on the Abalone 19 data. The AER-XGBoost with the exponential likelihood has a lower balanced accuracy, because of a relatively lower recall on minority instances; however, it is still higher than those of existing methods. On this dataset, the vanilla XGBoost, the focal loss neural network, and the LightGBM models all scramble to learn anything useful. The performances of the cost-sensitive SVM and decision tree are better than random guess, but still far from satisfactory. Once again, the focal-XGBoost exhibits a relatively competitive performance, although it is still inferior to the AER-XGBoost methods. It can be observed that the multiple entries in the table are given as '-'; this is because one or more metrics of the method is out of the ordinary range, which results in ridiculously high/low or even inf or Nan results. 

\begin{table}
\centering
\captionsetup{justification=centering}
\caption{\label{tab:mcnemar-abalone}McNemar's test for Log- and Exp-likelihood AERs against existing methods on the \textbf{Abalone 19 dataset}($\alpha=0.05$)}.
\begin{tabular}{|l|l|l|l|}
\hline
& $\chi^2$ statistics &  $p$-value & Null Hypothesis\\ \hhline{|=|=|=|=|}
\multicolumn{4}{|l|}{Log-likelihood AER} \\ \hline
Cost-sensitive Decision Tree & 112.23  & $<0.01$ & Reject $H_{0}$ \\ \hline
Cost-sensitive SVM & 11.68  & $<0.01$ & Reject $H_{0}$ \\ \hline
Focal loss neural network & 114.2  & $<0.01$ & Reject $H_{0}$ \\ \hline
LightGBM & 114.2  & $<0.01$ & Reject $H_{0}$ \\ \hline
Dynamic Ensemble (log) & 9.5 & $<0.01$ & Reject $H_{0}$ \\\hline
Plain XGBoost & 113.22 & $<0.01$ & Reject $H_{0}$ \\ \hline
Focal XGBoost & 0.68 & $0.41$ & Failed to reject $H_{0}$ \\ \hhline{|=|=|=|=|}
\multicolumn{4}{|l|}{Exp-likelihood AER} \\ \hline
Cost-sensitive Decision Tree & 75.26  & $<0.01$ &  Reject $H_{0}$\\ \hline
Cost-sensitive SVM & 0.08  & $0.77$ &  Failed to reject $H_{0}$\\ \hline
Focal loss neural network & 114.2  & $<0.01$ & Reject $H_{0}$ \\ \hline
LightGBM & 114.2  & $<0.01$ & Reject $H_{0}$ \\ \hline
Dynamic Ensemble (exp) & 4.3 & $0.04$ & Reject $H_{0}$ \\ \hline
Plain XGBoost & 4.0 & $<0.01$ &  Reject $H_{0}$\\ \hline
Focal XGBoost & 54.0 & $0.02$ & Reject $H_{0}$\\ \hline
\end{tabular}
\end{table}
\par

\begin{table}[h!]
\centering
\captionsetup{justification=centering}
\caption{\label{tab:Wilcoxon-abalone}Wilcoxon test for Log- and Exp-likelihood AERs against existing methods on the \textbf{Abalone 19 dataset}($\alpha=0.05$)}.
\begin{tabular}{|l|l|l|l|}
\hline
& $\chi^2$ statistics &  $p$-value & Null Hypothesis\\ \hhline{|=|=|=|=|}
\multicolumn{4}{|l|}{Log-likelihood AER} \\ \hline
Cost-sensitive Decision Tree & 274.0 & $<0.01$ & Reject $H_{0}$ \\ \hline
Cost-sensitive SVM &  4402.0 & $0.15$ & Failed to reject $H_{0}$ \\ \hline
Focal loss neural network & 0 & $<0.01$ & Reject $H_{0}$ \\ \hline
Light GBM & 0 & $<0.01$ & Reject $H_{0}$ \\ \hline
Dynamic Ensemble (log) & 136.5 & $<0.01$ & Reject $H_{0}$ \\ \hline
Plain XGBoost & 0 & $<0.01$ & Reject $H_{0}$ \\ \hline
Focal XGBoost & 3146.0 & $0.14$ & Failed to reject $H_{0}$ \\ \hhline{|=|=|=|=|}
\multicolumn{4}{|l|}{Exp-likelihood AER} \\ \hline
Cost-sensitive Decision Tree  & 188.0 & $<0.01$ &  Reject $H_{0}$\\ \hline
Cost-sensitive SVM  & 2331.0 & $0.01$ & Reject $H_{0}$\\ \hline
Focal loss neural network & 0 & $<0.01$ &  Reject $H_{0}$\\ \hline
LightGBM  & 0 & $<0.01$ &  Reject $H_{0}$\\ \hline
Dynamic Ensemble (exp) & 1064.0 & $0.03$ & Reject $H_{0}$ \\ \hline
Plain XGBoost & 0 & $<0.01$ &  Reject $H_{0}$\\ \hline
Focal XGBoost  & 3795.0 & $0.02$ & Reject $H_{0}$\\ \hline
\end{tabular}
\end{table}
\par

Another interesting perspective offered by the table is the comparison between the vanilla XGBoost (the base classifier used in the AER method) and the advanced methods based on it (including the focal loss proposed in \cite{wang2019imbalance} and the AER method proposed in this study). It can be observed that the plain XGBoost method performs poorly at this specific task, with a significant bias toward the minority data, and failure at spotting the majority instances. The focal loss and AER can be regarded as two ·approaches to improving the performance, and the AER is superior, in terms of the overall performance. \par

Finally, Tables \ref{tab:mcnemar-abalone} and \ref{tab:Wilcoxon-abalone} demonstrate the results of the Mcnemar's and Wilcoxon tests, to reveal the significance of the performance superiority of the AER models. From the tables, it can be observed that the preferable performances of the AER models are corroborated by both tests in most cases. The null hypothesis between the logarithm-based AER-XGBoost and the focal-loss XGBoost is relatively hard to reject ($p \geq 0.01$ for both tests), confirming the strong performance of the widely favored method. It is interesting that, for the SVM method, we failed to reject the null hypothesis for the exp-AER under the McNemar's test  and the log-AER under the Wilcoxon test. Such observations are not made elsewhere in the experiments. Thus, the problem in these specific datasets could stem from the specific training/testing split pairs.

\subsection{GMM-generated Data and Variations}
For the GMM-generated data, we first present the results on the $9$-center GMM-based set. For this model, the number of candidate Gaussian centroids is listed as $\{8,9,10\}$, and the $9$-centroid model is finally determined, which is consistent with the generative distribution of the data. The optimal $\lambda$ is determined as $0.60$ for the log-likelihood, and $0.30$ for the exp-likelihood. The validation set is selected to determine the value of $\delta$, and the performance, with respect to a varying $\delta$ value can be shown in Figures \ref{fig:GMMLogLikelihoodPerformance} and \ref{fig:GMMExpLikelihoodPerformance} for the log- and exp-likelihood methods respectively. \par
Similar to the previous experiments, the two types of the AER models and the methods of comparisons, including the cost-sensitive SVM and cost-sensitive decision tree, focal loss neural network, LightGBM, and the vanilla and focal-loss XGBoost models, are tested on the $9$-center GMM dataset. The results are summarized in Table \ref{tab:PerformanceComparisonGMMgenerate}. From the table, it can be found that only the cost-sensitive decision tree, focal-loss XGBoost, and the AER methods can grasp useful information, whereas other methods failed at learning more than some random classification boundaries. These results are consistent with what we have discussed earlier; the classification of the minority data is very hard to learn, because the 100 samples come from nine different Gaussian distributions. Nevertheless, with the AER methods, especially under the setup of the exp-likelihood, the model can maintain a relatively good performance. The results in Table \ref{tab:PerformanceComparisonGMMgenerate} indicates that if the data manifold really conforms to the GMM, the AER-XGBoost method can serve as a powerful learner that extracts information from the highly complicated manifold, where most of the other models will fail. \par
\begin{table}[h!]
\centering
\captionsetup{justification=centering}
\caption{\label{tab:PerformanceComparisonGMMgenerate}The comparison between the performance of different algorithms on the \textbf{9-center GMM-generated data}.}
\begin{tabular}{|p{4.5cm}|p{1.5cm}|p{1.5cm}|p{1.2cm}|p{1.0cm}|p{1.0cm}|p{1.5cm}|}
\hline
\abcdresults{ }{Balanced Accuracy}{Minority Recall}{Majority Recall}{F1 score}{G-mean}{TP-FP Ratio} \hline
\abcdresults{Cost-sensitive SVM}{50.00\%}{0\%}{100\%}{--}{--}{--}\hline
\abcdresults{Cost-sensitive Decision Tree }{52.09\%}{5.03\%}{99.18\%}{0.0588}{0.0598}{6.0769}\hline
\abcdresults{Focal loss neural network}{50.00\%}{0}{100\%}{--}{--}{--} \hline
\abcdresults{LightGBM}{50.00\%}{0}{100\%}{--}{--}{--} \hline
\abcdresults{Plain XGBoost}{50.03\%}{100\%}{0.063\%}{--}{--}{1.0006}\hline
\abcdresults{Focal-loss XGBoost}{54.74\%}{25.00\%}{84.49\%}{0.0370}{0.0707}{1.6122}\hline
\abcdresults{Dynamic Ensemble (log)}{48.35\% }{83.33\%}{0.47\%}{ 0.0132}{0.0195}{0.8373} \hline
\abcdresults{Dynamic Ensemble (exp)}{48.35\%}{83.33\%}{78.19\%}{ 0.0132}{0.0195}{3.8225} \hline
\abcdresults{AER-XGBoost (Log)}{\textbf{58.29\%}}{30.00\%}{86.58\%}{0.0504}{0.0909}{2.2358}\hline
\abcdresults{AER-XGBoost (Exp)}{57.53\%}{20.00\%}{95.06\%}{0.0743}{0.0988}{4.0513}\hline
\end{tabular}
\end{table}

The results of the hypothetical test on $9$-center GMM data, together with the performances and the testing results on other variations of the GMM-generated data, are given in Table \ref{tab:GMMInformationAll}. From the table, it can be observed that the AER method consistently outperforms all the compared methods, with a considerable margin. Moreover, both statistical hypothesis tests support the significance of the performance superiority. The exp-likelihood-based AER performs better on the hypothesis tests, indicating that this type of AER method can distinguish its predictions from that of other models more effectively.

\begin{table}[!htb]
\centering
\captionsetup{justification=centering}
\caption{\label{tab:GMMInformationAll}Results on the \textbf{GMM-generated datasets with different number of informative features}.}
\resizebox{\textwidth}{!}{
\begin{tabular}{|l|l|l|p{1.5cm}|p{2cm}|p{2cm}|p{2cm}|p{2cm}|}
\hline
& F1 Score & G-mean & Balanced Accuracy & (Wilcoxon)$H_{0}$ Log-AER-XGBoost & (Wilcoxon)$H_{0}$ Exp-AER-XGBoost & (Mcnemar)$H_{0}$ Log-AER-XGBoost & (Mcnemar)$H_{0}$ Exp-AER-XGBoost\\ \hhline{|=|=|=|=|=|=|=|=|}
\multicolumn{8}{|l|}{Informative Feature Number $n_{\text{informative}}=8$} \\ \hline
Cost-sensitive Decision Tree & 0.0816 & 0.0830 & 54.15\% & Reject & Reject & Reject & Reject \\ \hline
Cost-sensitive SVM & -- & -- & 50.00\% & Reject & Reject & Reject & Reject \\ \hline
Focal loss neural network & 0.005 & 0.0625 & 50.00\% & Reject & Reject & Reject & Reject \\ \hline
LightGBM & -- & -- & 50.00\% & Reject & Reject & Reject & Reject \\ \hline
Plain XGBoost & 0.0693 & 0.1160 & 61.96\% & Failed Reject & Reject & Failed Reject & Reject \\ \hline
Focal XGBoost & 0.0344 & 0.1186 & 61.74\% & Reject & Reject & Reject & Reject \\ \hline
Dynamic Ensemble (log) & 0.0653 & 0.1193 & 63.13 \% & Reject & -- & Reject & -- \\ \hline 
Dynamic Ensemble (exp) & 0.0653 & 0.1193 & 63.13 \% & -- & Reject & -- & Reject \\ \hline  
Log-AER-XGBoost & 0.0707 & 0.1173 & 62.09\% & -- & -- & -- & --\\ \hline
Exp-AER-XGBoost & 0.1159 & 0.1647 & \textbf{66.52\%} & -- & -- & -- & --\\ \hhline{|=|=|=|=|=|=|=|=|}
\multicolumn{8}{|l|}{Informative Feature Number $n_{\text{informative}}=9$} \\ \hline
Cost-sensitive Decision Tree & 0.0588 & 0.0598 & 52.09\% & Reject & Reject  & Reject & Reject \\ \hline
Cost-sensitive SVM & -- & -- & 50.00\% & Reject  & Reject &  Reject & Reject\\ \hline
Focal loss neural network & 0.0114 & 0.0115 & 50.00\% & Reject & Reject & Reject & Reject \\ \hline
LightGBM & -- & -- & 50.00\% & Reject & Reject & Reject & Reject \\ \hline
Plain XGBoost & -- & -- & 50.03\% & Reject & Reject & Reject & Reject\\ \hline
Focal XGBoost & 0.0370 & 0.0707 & 54.74\% & Reject & Reject & Failed Reject & Reject \\ \hline
Dynamic Ensemble (log) & 0.0132 & 0.0195 & 48.39 \% & Reject & -- & Reject & -- \\ \hline 
Dynamic Ensemble (exp) & 0.0132 & 0.0195 & 48.39 \% & -- & Reject & -- & Reject \\ \hline 
Log-AER-XGBoost & 0.0504 & 0.0909 & \textbf{58.29\%} & -- & -- & -- & --\\ \hline
Exp-AER-XGBoost & 0.0743 & 0.0988 & 57.53\% & -- & -- & -- & --\\ \hhline{|=|=|=|=|=|=|=|=|}
\multicolumn{8}{|l|}{Informative Feature Number $n_{\text{informative}}=10$} \\ \hline
Cost-sensitive Decision Tree & 0.1176 & 0.1195 & 54.62\% & Reject  & Reject & Reject & Reject \\ \hline
Cost-sensitive SVM & -- & -- & 50.00\% & Reject & Reject  & Reject & Reject \\ \hline
Focal loss neural network & 0.0179 & 0.0198 & 50.00\% & Reject & Reject & Reject & Reject \\ \hline
LightGBM & 0.0952 & 0.2236 & 50.00\% & Reject & Reject & Reject & Reject \\ \hline
Plain XGBoost & 0.0629 & 0.0948 & 58.26\% & Failed Reject & Reject  & Failed Reject & Reject \\ \hline
Focal XGBoost & 0.1091 & 0.1134 & 56.49\% & Reject & Failed Reject & Reject & Failed Reject \\ \hline
Dynamic Ensemble (log) & 0.0596 & 0.1067 & 60.92 \% & Reject & -- & Reject & -- \\ \hline 
Dynamic Ensemble (exp) & 0.0596 & 0.1067 & 60.92 \% & -- & Reject & -- & Reject \\ \hline 
Log-AER-XGBoost & 0.1032 & 0.1540 & 65.98\% & -- & -- & -- & --\\ \hline
Exp-AER-XGBoost & 0.1684 & 0.2066 & \textbf{67.88\%} & -- & -- & -- & --\\ \hhline{|=|=|=|=|=|=|=|=|}
\multicolumn{8}{|l|}{Informative Feature Number $n_{\text{informative}}=11$} \\ \hline
Cost-sensitive Decision Tree & -- & 0.0 & 49.36\% & Reject & Reject & Reject & Reject \\ \hline
Cost-sensitive SVM & -- & -- & 50.00\% & Reject & Reject & Reject & Reject \\ \hline
Focal loss neural network & 0.0146 & 0.0148 & 50.00\% & Reject & Reject & Reject & Reject \\ \hline
LightGBM & -- & -- & 50.00\% & Reject & Reject & Reject & Reject \\ \hline
Plain XGBoost & 0.0397 & 0.0586 & 53.44\% & Reject & Reject & Reject & Reject \\ \hline
Focal XGBoost & 0.0388 & 0.0656 & 54.24\% & Failed Reject & Reject & Failed Reject & Reject \\ \hline
Dynamic Ensemble (log) & 0.0690 & 0.1 & 58.70 \% & Reject & -- & Reject & -- \\ \hline 
Dynamic Ensemble (exp) & 0.0448 & 0.0785 & 56.23 \% & -- & Reject & -- & Reject \\ \hline 
Log-AER-XGBoost & 0.0863 & 0.1230 & 61.42\% & -- & -- & -- & --\\ \hline
Exp-AER-XGBoost & 0.1522 & 0.1845 & \textbf{65.44\%} & -- & -- & -- & --\\ \hhline{|=|=|=|=|=|=|=|=|}
\multicolumn{8}{|l|}{Informative Feature Number $n_{\text{informative}}=12$} \\ \hline
Cost-sensitive Decision Tree & -- & 0.00 & 49.43\% & Reject & Reject & Reject & Reject \\ \hline
Cost-sensitive SVM & -- & -- & 50.00\% & Reject & Reject & Reject & Reject \\ \hline
Cost-sensitive Network & 0.0128 & 0.0128 & 50.00\% & Reject & Reject & Reject & Reject \\ \hline
LightGBM & -- & -- & 50.00\% & Reject & Reject & Reject & Reject \\ \hline
Plain XGBoost & 0.0410 & 0.0747 & 55.57\% & Reject & Reject & Reject & Reject \\ \hline
Focal XGBoost & 0.0800 & 0.1000 & 52.37\% & Reject & Reject & Reject & Reject \\ \hline
Dynamic Ensemble (log) & 0.0339 & 0.0452 & 51.96 \% & Reject & -- & Failed Reject & -- \\ \hline 
Dynamic Ensemble (exp) & 0.0516 & 0.0770 & 55.85 \% &-- & Failed Reject & -- & Failed Reject \\ \hline 
Log-AER-XGBoost & 0.0662 & 0.0977 & 58.51\% & -- & -- & -- & --\\ \hline
Exp-AER-XGBoost & 0.1687 & 0.1972 & \textbf{65.72\%} & -- & -- & -- & --\\ \hhline{|-|-|-|-|-|-|-|-|}
\end{tabular}
}
\end{table}


\subsection{Other UCI Imbalanced Datasets}
Tables \ref{tab:PerformanceComparisonEcoli}-\ref{tab:Wilcoxon-car} show the results of performance and statistical hypothesis tests for five other UCI datasets, namely \textbf{Ecoli}, \textbf{US Crime}, \textbf{Wine Quality}, \textbf{Scene}, and \textbf{Car Eval}. The characteristics of these datasets have been discussed in Section \ref{subsec:data} and Table \ref{tab:uci-data-stat}. It can be observed that, compared with the three datasets we discussed earlier, these five datasets have considerably less imbalanced ratio. Therefore, the advantages of the AER-XGBoost model in this context are not as significant as in the above experiments. Nevertheless, they can still serve as powerful proofs of the effectiveness of the AER models. \par
Tables \ref{tab:PerformanceComparisonEcoli}-\ref{tab:Wilcoxon-ecoli} illustrate the performances and the results of the statistical tests on the Ecoli data. From the tables, it can be found that the AER model does not outperform the recent novel methods (e.g. Light GBM, XGBoost, and focal-XGBoost), although it is better than the conventional algorithms (e.g. SVM and decision tree). The inadequacy of the performance can be partially explained by the lower imbalance rate, which compromises the AER's specialization in processing imbalanced data, and prompts some recent general-purpose methods to unleash their abilities in handling ordinary datasets. Regardless of the performances in comparison to the other methods, the comparison with pure dynamic ensemble still validates the effectiveness of the AER framework. Furthermore, both types of statistical hypothesis tests suggest that the difference between the prediction of the AER-XGBoost and other methods is significant, to reject the null hypothesis in most of the cases.

\begin{table}[!htb]
\centering
\captionsetup{justification=centering}
\caption{\label{tab:PerformanceComparisonEcoli}The comparison between the performance of different algorithms on the \textbf{Ecoli data}.}
\begin{tabular}{|p{4.5cm}|p{1.5cm}|p{1.5cm}|p{1.2cm}|p{1.0cm}|p{1.0cm}|p{1.5cm}|}
\hline
\abcdresults{ }{Balanced Accuracy}{Minority Recall}{Majority Recall}{F1 score}{G-mean}{TP-FP Ratio} \hline
\abcdresults{Cost-sensitive SVM}{83.59\%}{75\%}{92.19\%}{0.5}{0.5303}{9.6}\hline
\abcdresults{Cost-sensitive Decision Tree}{74.22\%}{50\%}{98.44\%}{0.5714}{0.5774}{32.0}\hline
\abcdresults{Focal loss neural network}{50.00\%}{0}{100\%}{--}{--}{--}\hline
\abcdresults{LightGBM}{86.72\%}{75.00\%}{98.44\%}{0.75}{0.75}{48.0}\hline
\abcdresults{XGBoost}{92.19\%}{100\%}{84.38\%}{0.4444}{0.5345}{6.4}\hline
\abcdresults{Focal-loss XGBoost}{86.72\%}{75.00\%}{98.44\%}{0.75}{0.75}{48.0}\hline
\abcdresults{Dynamic Ensemble (Log)}{83.59\%}{75.00\%}{92.19\%}{0.5}{0.5303}{9.6}\hline
\abcdresults{Dynamic Ensemble (Exp)}{80.47\%}{75.00\%}{85.94\%}{0.375}{0.433}{5.3}\hline
\abcdresults{AER-XGBoost (Log)}{83.59\%}{75.00\%}{92.19\%}{0.5}{0.5303}{9.6}\hline
\abcdresults{AER-XGBoost (Exp)}{\textbf{84.38\%}}{75.00\%}{93.75\%}{0.5455}{0.5669}{12.0}\hline
\end{tabular}
\end{table}

\begin{table}[!htb]
\centering
\captionsetup{justification=centering}
\caption{\label{tab:mcnemar-ecoli}McNemar's test for Log- and Exp-likelihood AERs against existing methods on the \textbf{Ecoli dataset}($\alpha=0.05$)}.
\begin{tabular}{|l|l|l|l|}
\hline
& $\chi^2$ statistics & $p$-value & Null Hypothesis\\ \hhline{|=|=|=|=|}
\multicolumn{4}{|l|}{Log-likelihood AER} \\ \hline
Cost-sensitive Decision Tree & 4.17  & $<0.01$ & Reject $H_{0}$ \\ \hline
Cost-sensitive SVM & 0.5  & $0.48$ & Failed to Reject $H_{0}$ \\ \hline
Focal loss neural network & 4.17 & $<0.01$ & Reject $H_{0}$ \\ \hline
LightGBM & 4.17  & $<0.01$ & Reject $H_{0}$ \\ \hline
Plain XGBoost & 0.8 & $0.37$ & Failed to Reject $H_{0}$ \\ \hline
Focal XGBoost & 4.17 & $<0.01$ & Reject $H_{0}$ \\ \hline
Dynamic Ensemble (log) & 0.0 & 1.0 & Failed to Reject $H_{0}$ \\ \hhline{|=|=|=|=|}
\multicolumn{4}{|l|}{Exp-likelihood AER} \\ \hline
Cost-sensitive Decision Tree & 19.36  & $<0.01$ &  Reject $H_{0}$\\ \hline
Cost-sensitive SVM & 15.43 & $<0.01$ &  Reject $H_{0}$\\ \hline
Focal loss neural network & 4.17  & $<0.01$ & Reject $H_{0}$ \\ \hline
LightGBM & 4.17  & $<0.01$ & Reject $H_{0}$ \\ \hline
Plain XGBoost & 12.07 & $<0.01$ &  Reject $H_{0}$\\ \hline
Focal XGBoost & 19.36 & $<0.01$ & Reject $H_{0}$\\ \hline
Dynamic Ensemble (exp) & 11.53 & $<0.01$ & Reject $H_{0}$\\ \hline
\end{tabular}
\end{table}

\begin{table}[!htb]
\centering
\captionsetup{justification=centering}
\caption{\label{tab:Wilcoxon-ecoli}Wilcoxon test for Log- and Exp-likelihood AERs against existing methods on the \textbf{Ecoli dataset}($\alpha=0.05$)}.
\begin{tabular}{|l|l|l|l|}
\hline
& $\chi^2$ statistics &  $p$-value & Null Hypothesis\\ \hhline{|=|=|=|=|}
\multicolumn{4}{|l|}{Log-likelihood AER} \\ \hline
Cost-sensitive Decision Tree & 0 & $0.01$ & Reject $H_{0}$ \\ \hline
Cost-sensitive SVM &  0 & $0.16$ & Failed to reject $H_{0}$ \\ \hline
Focal loss neural network & 0 & $0.01$ & Reject $H_{0}$ \\ \hline
Light GBM & 0 & $0.01$ & Reject $H_{0}$ \\ \hline
Plain XGBoost & 0 & $0.02$ & Reject $H_{0}$ \\ \hline
Focal XGBoost & 0 & $0.01$ & Reject $H_{0}$ \\ \hline 
Dynamic Ensemble (log) & 6.0 & $0.65$ & Failed to reject $H_{0}$\\  \hhline{|=|=|=|=|}

\multicolumn{4}{|l|}{Exp-likelihood AER} \\ \hline
Cost-sensitive Decision Tree  & 0 & $<0.01$ &  Reject $H_{0}$\\ \hline
Cost-sensitive SVM  & 0 & $<0.01$ & Reject $H_{0}$\\ \hline
Focal loss neural network & 0 & $<0.01$ &  Reject $H_{0}$\\ \hline
Light GBM  & 0 & $<0.01$ &  Reject $H_{0}$\\ \hline
Plain XGBoost & 0 & $<0.01$ &  Reject $H_{0}$\\ \hline
Focal XGBoost  & 0 & $<0.01$ & Reject $H_{0}$\\ \hline 
Dynamic Ensemble (exp) & 0 & $<0.01$ & Reject $H_{0}$\\ \hline
\end{tabular}
\end{table}
\par

Tables \ref{tab:PerformanceComparisonUsCrime}-\ref{tab:Wilcoxon-uscrime} demonstrate the experimental results on the US Crime data. The performances in Table \ref{tab:PerformanceComparisonUsCrime} further demonstrate the advantages of the AER models, which outperform the classical models (SVM and decision tree) by a large margin, and achieve better scores than recent models such as the LightGBM, XGBoost and focal-XGBoost. The comparisons of the AER models with their pure likelihood-based (unregularized) counterparts also demonstrate the effectiveness of the regularization property. Furthermore, Tables \ref{tab:mcnemar-uscrime} and \ref{tab:Wilcoxon-uscrime} show the results of the hypothesis tests. Both tests suggest that the upper-hand performance superiority of the exponential-based AER-XGBoost is deemed significant. The logarithm-based AER-XGBoost performs relatively badly on the Mcnamar's test; however, the same results on the Wilcoxon test are mostly positive, suggesting that the failure of the Mcnamar's test can be attributed to the nature of the dataset.

\begin{table}[!htb]
\centering
\captionsetup{justification=centering}
\caption{\label{tab:PerformanceComparisonUsCrime}The comparison between the performance of different algorithms on the \textbf{US Crime data}.}
\begin{tabular}{|p{4.5cm}|p{1.5cm}|p{1.5cm}|p{1.2cm}|p{1.0cm}|p{1.0cm}|p{1.5cm}|}
\hline
\abcdresults{ }{Balanced Accuracy}{Minority Recall}{Majority Recall}{F1 score}{G-mean}{TP-FP Ratio} \hline
\abcdresults{Cost-sensitive SVM}{68.78\%}{42.86\%}{94.71\%}{0.36}{0.3647}{8.1}\hline
\abcdresults{Cost-sensitive Decision Tree}{69.84\%}{42.86\%}{96.83\%}{0.4286}{0.4286}{13.5}\hline
\abcdresults{Focal loss neural network}{50.00\%}{1.75\%}{98.25\%}{0.0263}{0.0304}{1.0}\hline
\abcdresults{LightGBM}{63.62\%}{28.57\%}{98.68\%}{0.3750}{0.3948}{21.60}\hline
\abcdresults{XGBoost}{75.93\%}{57.14\%}{94.71\%}{0.4528}{0.4629}{10.8}\hline
\abcdresults{Focal-loss XGBoost}{75.79\%}{57.14\%}{94.44\%}{0.4444}{0.4558}{10.3}\hline
\abcdresults{Dynamic Ensemble (Log)}{81.48\%}{71.43\%}{91.53\%}{0.4412}{0.4775}{8.4}\hline
\abcdresults{Dynamic Ensemble (Exp)}{87.96\%}{90.48\%}{85.45\%}{0.4}{0.4820}{6.2}\hline
\abcdresults{AER-XGBoost (Log)}{81.61\%}{71.437\%}{91.80\%}{0.4478}{0.4826}{8.7}\hline
\abcdresults{AER-XGBoost (Exp)}{\textbf{89.15\%}}{90.48\%}{87.83\%}{0.4419}{0.5143}{7.4}\hline
\end{tabular}
\end{table}

\begin{table}[!htb]
\centering
\captionsetup{justification=centering}
\caption{\label{tab:mcnemar-uscrime}McNemar's test for Log- and Exp-likelihood AERs against existing methods on the \textbf{US Crime dataset}($\alpha=0.05$)}.
\begin{tabular}{|l|l|l|l|}
\hline
& $\chi^2$ statistics &  $p$-value & Null Hypothesis\\ \hhline{|=|=|=|=|}
\multicolumn{4}{|l|}{Log-likelihood AER} \\ \hline
Cost-sensitive Decision Tree & 1.64 & $0.2$ & Failed to Reject $H_{0}$ \\ \hline
Cost-sensitive SVM & 0.59 & $0.44$ & Failed to Reject $H_{0}$ \\ \hline
Focal loss neural network & 6.92 & $<0.01$ & Reject $H_{0}$ \\ \hline
LightGBM & 6.92 & $<0.01$ & Reject $H_{0}$ \\ \hline
Plain XGBoost & 2.27 & $0.14$ & Failed to Reject $H_{0}$ \\ \hline
Focal XGBoost & 1.71 & $0.19$ & Failed to Reject $H_{0}$ \\\hline
Dynamic Ensemble (log) & 0.14 & 0.71 & Failed to Reject $H_{0}$ \\ \hhline{|=|=|=|=|}
\multicolumn{4}{|l|}{Exp-likelihood AER} \\ \hline
Cost-sensitive Decision Tree & 236.9 & $<0.01$ &  Reject $H_{0}$\\ \hline
Cost-sensitive SVM & 236.2 & $<0.01$ &  Reject $H_{0}$\\ \hline
Focal loss neural network & 0.64 & $0.42$ & Failed to Reject $H_{0}$ \\ \hline
LightGBM & 0.64 & $0.42$ & Failed to Reject $H_{0}$ \\ \hline
Plain XGBoost & 246.0 & $<0.01$ &  Reject $H_{0}$\\ \hline
Focal XGBoost & 245.0 & $<0.01$ & Reject $H_{0}$\\ \hline
Dynamic Ensemble (exp) & 241.3 & $<0.01$ & Reject $H_{0}$\\ \hline
\end{tabular}
\end{table}
\par

\begin{table}[!htb]
\centering
\captionsetup{justification=centering}
\caption{\label{tab:Wilcoxon-uscrime}Wilcoxon test for Log- and Exp-likelihood AERs against existing methods on the \textbf{US Crime dataset}($\alpha=0.05$)}.
\begin{tabular}{|l|l|l|l|}
\hline
& $\chi^2$ statistics &  $p$-value & Null Hypothesis\\ \hhline{|=|=|=|=|}
\multicolumn{4}{|l|}{Log-likelihood AER} \\ \hline
Cost-sensitive Decision Tree & 115.0 & $0.67$ & Failed to Reject $H_{0}$ \\ \hline
Cost-sensitive SVM &  47.5 & $0.06$ & Failed to Reject $H_{0}$ \\ \hline
Focal loss neural network & 7.5 & $<0.01$ & Reject $H_{0}$ \\ \hline
Light GBM & 7.5 & $<0.01$ & Reject $H_{0}$ \\ \hline
Plain XGBoost & 55.0 & $0.02$ & Reject $H_{0}$ \\ \hline
Focal XGBoost & 42.0 & $<0.01$ & Reject $H_{0}$ \\ \hline
Dynamic Ensemble (log) & 107.5 & $<0.01$ & Reject $H_{0}$ \\ \hhline{|=|=|=|=|}
\multicolumn{4}{|l|}{Exp-likelihood AER} \\ \hline
Cost-sensitiveDecision Tree  & 294.0 & $<0.01$ &  Reject $H_{0}$\\ \hline
Cost-sensitive SVM  & 0 & $<0.01$ & Reject $H_{0}$\\ \hline
Focal loss neural network & 0 & $<0.01$ &  Reject $H_{0}$\\ \hline
Light GBM  & 0 & $<0.01$ &  Reject $H_{0}$\\ \hline
Plain XGBoost & 0 & $<0.01$ &  Reject $H_{0}$\\ \hline
Focal XGBoost  & 0 & $<0.01$ & Reject $H_{0}$\\ \hline
Dynamic Ensemble (exp) & 0 & $<0.01$ & Reject $H_{0}$ \\ \hline
\end{tabular}
\end{table}
\par

The third dataset we tested the models on is the famous Wine Quality dataset, and the results are presented in Tables \ref{tab:PerformanceComparisonWineQuality}-\ref{tab:Wilcoxon-wine}. From Table \ref{tab:PerformanceComparisonWineQuality}, it can be observed that apart from the classical methods (SVM and decision tree), the performance of the AER is roughly in the same range as those of the other recent and advanced models. One exception is the focal-XGBoost, which scores $81.75\%$ in terms of balanced accuracy. However, this score can be attributed to the fact that it compromises the classification of the majority data, which also leads to lower F1 and G-mean scores. In addition to the performances, the results in Table \ref{tab:mcnemar-wine} suggest that the Mcnemar’s tests reject the Null hypothesis for all but three models, namely the cost-sensitive decision trees, focal loss neural network, and LightGBM. It may be observed that the results of the focal loss neural network and LightGBM are almost “all-majority” prediction; therefore, we can safely conclude that the failure of rejection is attributable to the imbalanced nature of the dataset. Furthermore, the results in Table \ref{tab:Wilcoxon-wine} suggest the Wilcoxon tests reject almost all the null hypotheses.

\begin{table}[!htb]
\centering
\captionsetup{justification=centering}
\caption{\label{tab:PerformanceComparisonWineQuality}The comparison between the performance of different algorithms on the \textbf{Wine Quality data}.}
\begin{tabular}{|p{4.5cm}|p{1.5cm}|p{1.5cm}|p{1.2cm}|p{1.0cm}|p{1.0cm}|p{1.5cm}|}
\hline
\abcdresults{ }{Balanced Accuracy}{Minority Recall}{Majority Recall}{F1 score}{G-mean}{TP-FP Ratio} \hline
\abcdresults{Cost-sensitive SVM}{70.63\%}{60\%}{81.26\%}{0.1593}{0.2347}{3.2}\hline
\abcdresults{Cost-sensitive Decision Tree}{60.82\%}{23.33\%}{98.32\%}{0.2642}{0.2665}{13.85}\hline
\abcdresults{Focal loss neural network}{50.00\%}{1.94\%}{98.06\%}{0.0237}{0.0244}{1.0}\hline
\abcdresults{LightGBM}{54.68\%}{10.00\%}{99.37\%}{0.1538}{0.1826}{15.83}\hline
\abcdresults{XGBoost}{76.11\%}{60\%}{92.21\%}{0.2951}{0.3426}{7.7}\hline
\abcdresults{Focal-loss XGBoost}{\textbf{81.75\%}}{76.67\%}{86.84\%}{0.2584}{0.3452}{5.83}\hline
\abcdresults{Dynamic Ensemble (Log)}{74.33\%}{56.67\%}{92.00\%}{0.2764}{0.3218}{7.1}\hline
\abcdresults{Dynamic Ensemble (Exp)}{74.21\%}{60\%}{88.42\%}{0.2278}{0.2904}{5.2}\hline
\abcdresults{AER-XGBoost (Log)}{74.28\%}{56.67\%}{91.89\%}{0.2742}{0.3201}{7.0}\hline
\abcdresults{AER-XGBoost (Exp)}{78.86\%}{66.67\%}{91.05\%}{0.2963}{0.3563}{7.5}\hline
\end{tabular}
\end{table}

\begin{table}[!htb]
\centering
\captionsetup{justification=centering}
\caption{\label{tab:mcnemar-wine}McNemar's test for Log- and Exp-likelihood AERs against existing methods on the \textbf{Wine Quality dataset}($\alpha=0.05$)}.
\begin{tabular}{|l|l|l|l|}
\hline
& $\chi^2$ statistics &  $p$-value & Null Hypothesis\\ \hhline{|=|=|=|=|}
\multicolumn{4}{|l|}{Log-likelihood AER} \\ \hline
Cost-sensitive Decision Tree & 0.13  & $0.72$ & Failed to Reject $H_{0}$ \\ \hline
Cost-sensitive SVM & 133.0  & $<0.01$ & Reject $H_{0}$ \\ \hline
Focal loss neural network & 0.24  & $0.63$ & Failed to Reject $H_{0}$ \\ \hline
LightGBM & 0.24  & 0.63 & Failed to Reject $H_{0}$ \\ \hline
Plain XGBoost & 33.3 & $<0.01$ & Reject $H_{0}$ \\ \hline
Focal XGBoost & 68.4 & $<0.01$ & Reject $H_{0}$ \\ \hline
Dynamic Ensemble (log) & 5.6 & $0.01$ & Reject $H_{0}$ \\\hhline{|=|=|=|=|}
\multicolumn{4}{|l|}{Exp-likelihood AER} \\ \hline
Cost-sensitive Decision Tree & 3.52  & 0.06 &  Failed to Reject $H_{0}$\\ \hline
Cost-sensitive SVM & 116.3  & $<0.01$ &  Reject $H_{0}$\\ \hline
Focal loss neural network & 0.24  & $0.63$ & Failed to Reject $H_{0}$ \\ \hline
LightGBM & 0.24  & 0.63 & Failed to Reject $H_{0}$ \\ \hline
Plain XGBoost & 14.0 & $<0.01$ &  Reject $H_{0}$\\ \hline
Focal XGBoost & 53.8  & $<0.01$ & Reject $H_{0}$\\ \hline
Dynamic Ensemble (exp) & 55.7 & $<0.01$ & Reject $H_{0}$ \\\hline
\end{tabular}
\end{table}
\par

\begin{table}[!htb]
\centering
\captionsetup{justification=centering}
\caption{\label{tab:Wilcoxon-wine}Wilcoxon test for Log- and Exp-likelihood AERs against existing methods on the \textbf{Wine Quality dataset}($\alpha=0.05$)}.
\begin{tabular}{|l|l|l|l|}
\hline
& $\chi^2$ statistics &  $p$-value & Null Hypothesis\\ \hhline{|=|=|=|=|}
\multicolumn{4}{|l|}{Log-likelihood AER} \\ \hline
Cost-sensitive Decision Tree & 240.0 & $0.86$ & Failed to Reject $H_{0}$ \\ \hline
Cost-sensitive SVM &  88.5 & $<0.01$ & Reject $H_{0}$ \\ \hline
Focal loss neural network & 18.0 & $<0.01$ & Reject $H_{0}$ \\ \hline
Light GBM & 18.0 & $<0.01$ & Reject $H_{0}$ \\ \hline
Plain XGBoost & 36. & $<0.01$ & Reject $H_{0}$ \\ \hline
Focal XGBoost & 199.5 & $<0.01$ & Reject $H_{0}$ \\ \hline
Dynamic Ensemble (log) & 126.0 & $<0.01$ & Reject $H_{0}$ \\ \hhline{|=|=|=|=|}
\multicolumn{4}{|l|}{Exp-likelihood AER} \\ \hline
Cost-sensitive Decision Tree  & 318.5 & $<0.01$ &  Reject $H_{0}$\\ \hline
Cost-sensitive SVM  & 320.0 & $<0.01$ & Reject $H_{0}$\\ \hline
Focal loss neural network & 19.5 & $<0.01$ &  Reject $H_{0}$\\ \hline
Light GBM  & 19.5 & $<0.01$ &  Reject $H_{0}$\\ \hline
Plain XGBoost & 481.0 & $<0.01$ &  Reject $H_{0}$\\ \hline
Focal XGBoost  & 285.0 & $<0.01$ & Reject $H_{0}$\\ \hline
Dynamic Ensemble(exp) & 0 & $<0.01$ & Reject $H_{0}$ \\\hline
\end{tabular}
\end{table}
\par

Th Scene dataset, as mentioned earlier, has richer information, with a greater number of features. The results in Table \ref{tab:PerformanceComparisonScene} reveal that the advantages of the AER model are more compelling in this type of scenario. We can observe from the table that the performance of AER-XGBoosts is significantly better than that of the other models, including the classical ones and the recently developed ones. Once more, the vanilla and focal XGBoost models yield relatively more satisfactory performances, as they are the only models that can compete against the AER models. The improvements in the performance of the AER over the pure dynamic ensemble further demonstrates the positive impact of regularization on the dynamic ensemble, especially when the data information is rich. Additionally, the results in Tables \ref{tab:mcnemar-scene} and \ref{tab:Wilcoxon-scene} give valid statistical testing results.

\begin{table}[!htb]
\centering
\captionsetup{justification=centering}
\caption{\label{tab:PerformanceComparisonScene}The comparison between the performance of different algorithms on the \textbf{Scene data}.}
\begin{tabular}{|p{4.5cm}|p{1.5cm}|p{1.5cm}|p{1.2cm}|p{1.0cm}|p{1.0cm}|p{1.5cm}|}
\hline
\abcdresults{ }{Balanced Accuracy}{Minority Recall}{Majority Recall}{F1 score}{G-mean}{TP-FP Ratio} \hline
\abcdresults{Cost-sensitive SVM}{61.45\%}{26.67\%}{96.24\%}{0.2909}{0.2921}{7.1}\hline
\abcdresults{Cost-sensitive Decision Tree}{55.80\%}{20.00\%}{91.59\%}{0.1622}{0.1651}{2.38}\hline
\abcdresults{Focal loss neural network}{50.00\%}{0\%}{100\%}{--}{--}{--}\hline
\abcdresults{LightGBM}{51.67\%}{3.33\%}{100\%}{0.0645}{0.1826}{--}\hline
\abcdresults{XGBoost}{69.51\%}{66.67\%}{72.35\%}{0.2286}{0.3032}{2.4}\hline
\abcdresults{Focal-loss XGBoost}{66.95\%}{66.67\%}{73.23\%}{0.2339}{0.3075}{2.5}\hline
\abcdresults{Dynamic Ensemble (Log)}{67.35\%}{43.33\%}{91.37\%}{0.3171}{0.3291}{5.02}\hline
\abcdresults{Dynamic Ensemble (Exp)}{62.24\%}{83.33\%}{41.15\%}{0.1558}{0.2676}{1.4}\hline
\abcdresults{AER-XGBoost (Log)}{\textbf{71.58\%}}{53.33\%}{89.82\%}{0.3478}{0.3710}{5.2}\hline
\abcdresults{AER-XGBoost (Exp)}{71.50\%}{66.67\%}{76.33\%}{0.2548}{0.3240}{2.8}\hline
\end{tabular}
\end{table}

\begin{table}[!htb]
\centering
\captionsetup{justification=centering}
\caption{\label{tab:mcnemar-scene}McNemar's test for Log- and Exp-likelihood AERs against existing methods on the \textbf{Scene dataset}($\alpha=0.05$)}.
\begin{tabular}{|l|l|l|l|}
\hline
& $\chi^2$ statistics &  $p$-value & Null Hypothesis\\ \hhline{|=|=|=|=|}
\multicolumn{4}{|l|}{Log-likelihood AER} \\ \hline
Cost-sensitive Decision Tree & 96.9 & $<0.01$ & Reject $H_{0}$ \\ \hline
Cost-sensitive SVM & 114.3  & $<0.01$ & Reject $H_{0}$ \\ \hline
Focal loss neural network & 116.8  & $<0.01$ & Reject $H_{0}$ \\ \hline
LightGBM & 116.8  & $<0.01$ & Reject $H_{0}$ \\ \hline
Plain XGBoost & 16.5 & $<0.01$ & Reject $H_{0}$ \\ \hline
Focal XGBoost & 61.8 & $<0.01$ & Reject $H_{0}$ \\ \hline
Dynamic Ensemble (log) & 0.5 & $0.46$ & Failed to Reject $H_{0}$ \\\hhline{|=|=|=|=|}
\multicolumn{4}{|l|}{Exp-likelihood AER} \\ \hline
Cost-sensitive Decision Tree & 242.3  & $<0.01$ &  Reject $H_{0}$\\ \hline
Cost-sensitive SVM & 228.0  & $<0.01$ &  Reject $H_{0}$\\ \hline
Focal loss neural network & 116.8  & $<0.01$ & Reject $H_{0}$ \\ \hline
LightGBM & 116.8  & $<0.01$ & Reject $H_{0}$ \\ \hline
Plain XGBoost & 151.2  & $<0.01$ &  Reject $H_{0}$\\ \hline
Focal XGBoost & 119.7  & $<0.01$ & Reject $H_{0}$\\ \hline
Dynamic Ensemble (exp) & 32.6 & $<0.01$ & Reject $H_{0}$ \\\hline
\end{tabular}
\end{table}
\par

\begin{table}[!htb]
\centering
\captionsetup{justification=centering}
\caption{\label{tab:Wilcoxon-scene}Wilcoxon test for Log- and Exp-likelihood AERs against existing methods on \textbf{Scene dataset}($\alpha=0.05$)}.
\begin{tabular}{|l|l|l|l|}
\hline
& $\chi^2$ statistics &  $p$-value & Null Hypothesis\\ \hhline{|=|=|=|=|}
\multicolumn{4}{|l|}{Log-likelihood AER} \\ \hline
Cost-sensitive Decision Tree & 756.0 & $<0.01$ & Reject $H_{0}$ \\ \hline
Cost-sensitive SVM &  227.5 & $<0.01$ & Reject $H_{0}$ \\ \hline
Focal loss neural network & 0 & $<0.01$ & Reject $H_{0}$ \\ \hline
Light GBM & 0 & $<0.01$ & Reject $H_{0}$ \\ \hline
Plain XGBoost & 1274.0 & $<0.01$ & Reject $H_{0}$ \\ \hline
Focal XGBoost & 1008.0 & $<0.01$ & Reject $H_{0}$ \\ \hline
Dynamic Ensemble (log) & 1820.0 & $<0.29$ & Failed to Reject $H_{0}$ \\\hhline{|=|=|=|=|}
\multicolumn{4}{|l|}{Exp-likelihood AER} \\ \hline
Cost-sensitive Decision Tree  & 474.0 & $<0.01$ &  Reject $H_{0}$\\ \hline
Cost-sensitive SVM  & 320.0 & $<0.01$ & Reject $H_{0}$\\ \hline
Focal loss neural network & 0 & $<0.01$ &  Reject $H_{0}$\\ \hline
Light GBM  & 0 & $<0.01$ &  Reject $H_{0}$\\ \hline
Plain XGBoost & 796.0 & $<0.01$ &  Reject $H_{0}$\\ \hline
Focal XGBoost  & 388.5 & $<0.01$ & Reject $H_{0}$\\ \hline
Dynamic Ensemble(exp) & 264.0 & $<0.01$ & Reject $H_{0}$ \\\hline
\end{tabular}
\end{table}
\par

Finally, Tables \ref{tab:PerformanceComparisonCar}-\ref{tab:Wilcoxon-car} present the performances and results of the statistical tests on the Car Eval dataset. It can be observed that, as the dataset is relatively easy to learn (small size and low imbalance rate), almost all the models can yield good performance. Nevertheless, the AER-based models still yield better overall performances. It is also noticeable that, because the dataset is small, the dynamic ensemble without regularization can only offer performances comparable with classical methods, and the performances increase drastically after the introduction of regularization. In addition, the results of the hypothesis tests in Tables \ref{tab:mcnemar-car} and \ref{tab:Wilcoxon-car} suggest the differences in the predictions are significant.

\begin{table}[!htb]
\centering
\captionsetup{justification=centering}
\caption{\label{tab:PerformanceComparisonCar}Comparison between the performance of different algorithms on the \textbf{Car Eval data}.}
\begin{tabular}{|p{4.5cm}|p{1.5cm}|p{1.5cm}|p{1.2cm}|p{1.0cm}|p{1.0cm}|p{1.5cm}|}
\hline
\abcdresults{ }{Balanced Accuracy}{Minority Recall}{Majority Recall}{F1 score}{G-mean}{TP-FP Ratio} \hline
\abcdresults{Cost-sensitive SVM}{93.38\%}{89.29\%}{97.48\%}{0.8197}{0.8224}{35.5}\hline
\abcdresults{Cost-sensitive Decision Tree}{88.71\%}{82.14\%}{95.28\%}{0.6970}{0.7051}{17.4}\hline
\abcdresults{Focal loss neural network}{50.00\%}{0}{100\%}{--}{--}{--}\hline
\abcdresults{LightGBM}{95.01\%}{92.86\%}{97.17\%}{0.8254}{0.8305}{32.8}\hline
\abcdresults{XGBoost}{94.34\%}{100\%}{88.68\%}{0.6087}{0.6614}{8.8}\hline
\abcdresults{Focal-loss XGBoost}{93.71\%}{100\%}{87.42\%}{0.5833}{0.6417}{8.0}\hline
\abcdresults{Dynamic Ensemble (Log)}{96.80\%}{96.43\%}{97.17\%}{0.8438}{0.8504}{34.1}\hline
\abcdresults{Dynamic Ensemble (Exp)}{94.38\%}{92.86\%}{95.91\%}{0.7761}{0.7868}{22.7}\hline
\abcdresults{AER-XGBoost (Log)}{95.01\%}{92.88\%}{97.17\%}{0.8254}{0.8305}{32.8}\hline
\abcdresults{AER-XGBoost (Exp)}{\textbf{97.96\%}}{100\%}{95.91\%}{0.8116}{0.8264}{24.5}\hline
\end{tabular}
\end{table}

\begin{table}[!htb]
\centering
\captionsetup{justification=centering}
\caption{\label{tab:mcnemar-car}McNemar's test for Log- and Exp-likelihood AERs against existing methods on the \textbf{Car Eval dataset}($\alpha=0.05$)}.
\begin{tabular}{|l|l|l|l|}
\hline
& $\chi^2$ statistics &  $p$-value & Null Hypothesis\\ \hhline{|=|=|=|=|}
\multicolumn{4}{|l|}{Log-likelihood AER} \\ \hline
Cost-sensitive Decision Tree & 4.1  & $<0.01$ & Reject $H_{0}$ \\ \hline
Cost-sensitive SVM & 13.0  & $<0.01$ & Reject $H_{0}$ \\ \hline
Focal loss neural network & 14.1  & $<0.01$ & Reject $H_{0}$ \\ \hline
LightGBM & 14.1  & $<0.01$ & Reject $H_{0}$ \\ \hline
Plain XGBoost & 1.8 & $0.18$ & Failed to Reject $H_{0}$ \\ \hline
Focal XGBoost & 3.4 & $0.07$ & Failed to Reject $H_{0}$ \\ \hline
Dynamic Ensemble(log) & 8.5 & $<0.01$ & Reject $H_{0}$ \\\hhline{|=|=|=|=|}
\multicolumn{4}{|l|}{Exp-likelihood AER} \\ \hline
Cost-sensitive Decision Tree & 325.1  & $<0.01$ &  Reject $H_{0}$\\ \hline
Cost-sensitive SVM & 312.2  & $<0.01$ &  Reject $H_{0}$\\ \hline
Focal loss neural network & 14.1  & $<0.01$ & Reject $H_{0}$ \\ \hline
LightGBM & 14.1  & $<0.01$ & Reject $H_{0}$ \\ \hline
Plain XGBoost & 304.0  & $<0.01$ &  Reject $H_{0}$\\ \hline
Focal XGBoost & 300.0  & $<0.01$ & Reject $H_{0}$\\ \hline
Dynamic Ensemble(exp) & 321.1 & $<0.01$ & Reject $H_{0}$ \\\hline
\end{tabular}
\end{table}
\par

\begin{table}[!htb]
\centering
\captionsetup{justification=centering}
\caption{\label{tab:Wilcoxon-car}Wilcoxon test for Log- and Exp-likelihood AERs against existing methods on the \textbf{Car Eval dataset}($\alpha=0.05$)}.
\begin{tabular}{|l|l|l|l|}
\hline
& $\chi^2$ statistics &  $p$-value & Null Hypothesis\\ \hhline{|=|=|=|=|}
\multicolumn{4}{|l|}{Log-likelihood AER} \\ \hline
Cost-sensitive Decision Tree & 0 & $<0.01$ & Reject $H_{0}$ \\ \hline
Cost-sensitive SVM &  0 & $<0.01$ & Reject $H_{0}$ \\ \hline
Focal loss neural network & 0 & $<0.01$ & Reject $H_{0}$ \\ \hline
Light GBM & 0 & $<0.01$ & Reject $H_{0}$ \\ \hline
Plain XGBoost & 30.0 & $0.11$ & Failed to Reject $H_{0}$ \\ \hline
Focal XGBoost & 87.5 & $<0.01$ & Reject $H_{0}$ \\ \hline
Dynamic Ensemble(exp) & 0 & $<0.01$ & Reject $H_{0}$ \\\hhline{|=|=|=|=|}
\multicolumn{4}{|l|}{Exp-likelihood AER} \\ \hline
Cost-sensitive Decision Tree  & 0 & $<0.01$ &  Reject $H_{0}$\\ \hline
Cost-sensitive SVM  & 0 & $<0.01$ & Reject $H_{0}$\\ \hline
Focal loss neural network & 0 & $<0.01$ &  Reject $H_{0}$\\ \hline
Light GBM  & 0 & $<0.01$ &  Reject $H_{0}$\\ \hline
Plain XGBoost & 0 & $<0.01$ &  Reject $H_{0}$\\ \hline
Focal XGBoost  & 0 & $<0.01$ & Reject $H_{0}$\\ \hline
Dynamic Ensemble(exp) & 0 & $<0.01$ & Reject $H_{0}$ \\\hline
\end{tabular}
\end{table}

\FloatBarrier

\section{Discussions}
\label{sec:discussion}
We dedicate this section to discussing the implications of the foregoing theoretical and empirical analysis and the missing details in the experiments. Specifically, we want to discuss the following aspects: 1. The effectiveness of the regularization; 2. The suitable problem for the AER and the choice between logarithm- and exponential-based AERs; 3. The practical training time and training dynamics of the AER; and 4. Natural improvements and extensions of the AER. 

From the experiments in Section \ref{ExperimentSection}, it can be observed that the AER-XGBoost models (with regularization) almost always outperform their pure dynamic ensemble counterparts in terms of the balanced accuracy. The only notable exceptions are in Tables \ref{tab:PerformanceComparisonAbalone} and \ref{tab:PerformanceComparisonGMMgenerate}, where the unregularized exp-based dynamic ensemble models achieve much better balanced accuracy. However, this “performance flux” is at the cost of the majority recall, which results in low TP-FP ratios, and renders the models practically inapplicable. Furthermore, for most of the statistical hypothesis tests, the regularized AER-XGBoosts reject the prediction null hypothesis from their pure dynamic ensemble counterparts. For the UCI Bioassay data, from Figures \ref{fig:WeightInterpolationLog} and \ref{fig:WeightInterpolationExp}, it can be observed that the optimal $\lambda$ for the validation and testing sets is neither $0$ nor $1$, indicating that interpolating the likelihood-based weights resulted in performance gain. The results on Abalone 19 and the GMM-generated data reveal similar results, which can further support the above observation. 

As we have discussed based on the experiments, the AER is best suitable for problems with complex decision boundary and high imbalanced ratio. For instance, the performance of the AER-XGBoost only offers competitive performances on the Ecoli dataset (imbalance ratio $1:8.6$), whereas it offers significantly state-of-the-art performances on the Abalone 19 data (imbalance ratio $1:129$). Furthermore, from the observations, based on most of the experiments, the exponential likelihood-based AERs are often the better ones. We believe this may be attributable to the fact that they can benefit more from regularization. This can be explained by the fact that the logarithm-based AERs have weights closer to uniform, whereas the weights of the exponential-based AERs distribute more “sharply.” The major drawback of the exponential-based AER, compared with the logarithm-based one, is that the range of choice for the “good” hyper-parameters is smaller. Thus, we recommend using the logarithm-based AER models when we have no knowledge of the relations between the training/validation and the test data, and using the exponential-based AER when we know the data distribution between the training/validation and the test sets are similar.

We theoretically proved in Section \ref{ComplexitySection} that the time and memory complexity of the models implemented under the AER framework is better than the vanilla framework. In practice, we observe that the running time for the AER-XGBoost is usually longer than expected. This can be explained by two factors. The first factor is that the time complexity of the XGBoost does not fall into the regime of polynomial time. Because the model includes parallelization during fitting, the complexity introduced by the number of data becomes $O(\log(m))$ (polylog) instead. In practice, we do not expect this type of classifier very often, and if the parallelization technique is enabled, we can also bring the GMM to parallel computing, to further improve the time complexity. The second factor is that we do not implement diagonal approximations for the co-variance matrices in the GMM models. Therefore, it takes $O(n^{3})$ complexity to compute the matrix inverse and multiplications (in contrast with the $O(n)$ complexity using diagonal approximation). We employed full co-variance matrices to illustrate the power of the AER model, and for real-life applications, it is recommended that the diagonal approximation is used when the dimension (number of features) of the data is large.

Some natural improvements and extensions will easily follow, if we apply the AER model, with augmentation by other methods. For instance, we can change the distribution from the GMM to a mixture of t-distributions, to obtain more robust modeling for the data \cite{peel2000robust}. The major disadvantage of this strategy is the time complexity may change. In addition, some compression methods, such as quantization \cite{ruisen2019pami}, can help reduce the memory complexity on the bit-level. It can also be observed that, because the AER is based on the GMM model, good initialization will contribute to better performances. Algorithms, such as K-means++ \cite{ostrovsky2013effectiveness}, are useful initializations on individual Gaussian distributions, whereas some quantization-based methods, such as the scalar quantization in \cite{ruisen2019pami}, can serve as convenient methods to provide initial separations for different Gaussian clusters.

\section{Conclusion}
\label{ConclusionSection}
In this paper, a novel method, the adaptive ensemble of classifiers with regularization (AER), has been proposed for binary imbalanced data classification. The details of the method, including implementations with the XGBoost, are provided, and related training formulas are derived. In addition to the regularization properties, we illustrate that the method has favorable time and memory complexity. The performance of the proposed algorithm is tested on multiple datasets, and empirical evidences illustrate that the performances are better or competitive, compared to the classical algorithms and the recent ones. In addition, the proposed method has advantages in terms of dealing with \emph{highly} imbalanced data, and benefits the research on solving overfitting problem, especially for dynamic ensemble methods.

Three major contributions have been made in this study. Firstly, we propose an algorithm with state-of-the-art performance on binary imbalanced data. Compared with the existing optimization methods and recent developments in the area (such as focal loss and LightGBM), the performance of the proposed method is, at least, competitive, in terms of multiple metrics. Secondly, the proposed method has multiple advantages other than classification performance, including better time and memory complexity and a flexible framework compatible with a majority of binary classifiers. Finally, we have investigated the regularization problem in dynamic ensemble methods, which is relatively underexplored in the previous publications. Experimental results show that the regularization techniques improve the performances of various models; furthermore, they have excellent potentials with regard to the classification of multi-class imbalanced data, e.g., one can decompose the problem of multi-class imbalanced data classification into a series of jobs of binary imbalanced data classification.

In the future, we plan to further explore the AER model from both the method and the application perspectives. On the method level, we can extend the AER model to multi-class imbalanced classification by combining binary AERs; Alternatively, we can study more advanced dynamic ensemble methods to represent multiple classes. The exploration of other forms of generalization is also a promising direction. From the application perspective, we are interested in applying the proposed AER method to various real-life problems with imbalanced data. Since the label-imbalance of datasets exists pervasively in industrial practice, there should be a wide range of applications for the AER method.

\section*{Acknowledgement}
We appreciate the earlier contributions from Qin Yu, Hang Zhang, Yanmei Yu, and Chao Sun to the paper. We also thank Michael Tan of University College London for his writing suggestions.

\section*{Funding Statement}
This work is supported by the Sichuan Science and Technology Program (2020YFG0051), and the University-Enterprise Cooperation Projects (17H1199, 19H0355, 19H1121).


\bibliographystyle{unsrt}
\bibliography{sample}

\clearpage
\appendix
\section{Hyper-parameters for Experiments}
In this section, we provide hyper-parameter sets for experiments in section \ref{ExperimentSection} for reproducibility, demonstrated in table \ref{tab:params}. Note that the parameters not mentioned are set to the default value of the corresponding models. \par

\begin{table}[!htb]
\centering
\captionsetup{justification=centering}
\caption{\label{tab:params}Parmaeters Applied to All Methods}.
\begin{tabular}{|l|l|}
\hline
Method & Parameters  \\ \hhline{|=|=|}
Cost-sensitive Decision Tree & \makecell[l]{$\text{tuned\_params}=\{"\text{class\_weight}":[\{0: 1, 1: 110\}, \{0: 1, 1: 130\}$, \\  $\{0: 1, 1: 150\}, 'balanced'],
               "\text{criterion}": ['gini', 'entropy']\}$}  \\ \hline
Cost-sensitive SVM & \makecell*[l]{$\text{tuned\_params}=\{"\text{class\_weight}":[\{0: 1, 1: 110\}, \{0: 1, 1: 130\}$,\\ $\{0: 1, 1: 150\}, 'balanced'],"\text{kernel}":['linear', 'rbf', 'poly']$,\\ $"\text{degree}":[3]\}$  }  \\ \hline
Focal loss neural network & $\alpha=0.25, \gamma=3$ \\ \hline
LightGBM & \makecell*[l]{$\text{tuned\_params}=\{"\text{eta}":[0.1,0.3,0.5,0.7],"\text{max\_depth}":[5,6,7,8,9,10]$,\\ $"\text{num\_round}":[7,8,9,10]\}$}   \\ \hline

Focal XGBoost & $\text{tuned\_params}=\{ \text{focal\_gamma}=[1.0,1.5,2.0]\}$  \\ \hline
Dynamic Ensemble & \makecell*[l]{$\text{tuned\_params}=\{"\text{n\_estimators}":[40,45,50,55,60],"\text{max\_depth}":[6,7,8,9]$,\\ $"\text{num\_leaves}":[20,30,40,50,60]\}$ 
\\ $\text{model\_num} = [9,10,11,12,13,14]$ } \\ \hline
AER & \makecell*[l]{$\text{tuned\_params}=\{"\text{eta}":[0.1,0.3,0.5,0.7],"\text{max\_depth}":[5,6,7,8,9,10]$,\\ $"\text{num\_round}":[7,8,9,10]\}$ 
\\ $\text{model\_num} = [9,10,11,12,13,14]$ } \\ \hline
\end{tabular}
\end{table}
\par

\end{document}